\documentclass[fleqn,10pt]{SelfArx} 

\usepackage{lipsum} 

\definecolor{color1}{RGB}{0,0,90} 
\definecolor{color2}{RGB}{0,20,20} 

\usepackage{hyperref} 
\hypersetup{hidelinks,colorlinks,breaklinks=true,urlcolor=color2,citecolor=color1,linkcolor=color1,bookmarksopen=false,pdftitle={Title},pdfauthor={Author}}

\JournalInfo{Draft Version} 
\Archive{} 

\PaperTitle{Comparative Analysis of Polygon-Based and Global Machine Learning Models for Bus Occupancy Prediction} 

\Authors{ Daniel Azenkot\textsuperscript{1}*, Eran Ben Elia\textsuperscript{2}, Michael Fire\textsuperscript{1}} 
\affiliation{\textsuperscript{1}\textit{Faculty of Computer and Information Science, University of Ben Gurion, Be'er Sheva, Israel}} 
\affiliation{\textsuperscript{2}\textit{Department of Geography and Environmental Development, Ben-Gurion, Israel}} 

\affiliation{*\textbf{Corresponding authors}: azendan@bgu.ac.il, benelia@bgu.ac.il, mickyfi@bgu.ac.il} 

\newcommand{\keywordname}{Keywords} 

\usepackage{hyperref} 
\hypersetup{hidelinks,colorlinks,breaklinks=true,urlcolor=color2,citecolor=color1,linkcolor=color1,bookmarksopen=false,pdftitle={Title},pdfauthor={Author}}
\usepackage[T1]{fontenc}
\usepackage[utf8]{inputenc}
\usepackage{graphicx}
\usepackage[numbers,square]{natbib}
\usepackage{color, soul}
\usepackage{comment}
\usepackage{pdfpages}
\usepackage{longtable}
\usepackage{float}
\usepackage{placeins}
\usepackage{amsmath}
\usepackage{tabularx}
\usepackage{lscape}
\usepackage{rotating}
\usepackage{booktabs}
\usepackage{rotating}
\usepackage{xtab}

\usepackage{subcaption}
\usepackage{siunitx}
\usepackage[toc,page]{appendix}
\usepackage{adjustbox} 

\JournalInfo{Draft Version} 

\Abstract{Accurate forecasting of bus ridership (passengers numbers) is crucial for efficient management and optimization of public transport systems. Traditional forecasting models often fail to capture the unique and localized dynamics of different urban areas by treating the entire city as a single, homogeneous region. This paper introduces a novel framework that enhances bus ridership prediction by integrating a spatial clustering methodology with multi-dimensional feature analysis.
The proposed framework utilizes a diverse set of data, including bus ridership data (by route number, time, and bus stop) complemented by a variety of open source data, such as spatial features (e.g., attractive destinations), meteorological conditions (e.g., temperature, rainfall), and temporal patterns (e.g., time of day, day of week). By clustering the urban area into distinct regions, based on the principle that bus stops in close proximity share similar ridership characteristics, a separate local forecasting model is trained for each of these clusters. This localized approach demonstrates an accuracy comparable to that of global models. The findings suggest that a spatially-aware, localized modeling strategy is effective for public transport prediction, paving the way for more targeted and efficient service improvements.}

\begin{document}

\flushbottom 

\maketitle 


\thispagestyle{empty} 


\section{Introduction} 
\addcontentsline{toc}{section}{Introduction} 

Public transport (PT) plays a vital role in the daily lives of people in many urban areas. During the past century, the global population has increasingly concentrated in cities, increasing the dependence on PT for everyday mobility ~\cite{petrovic2016appraisal}. PT provides regularly scheduled services accessible to all paying passengers, accommodating various trips with varying origins, destinations, and purposes. Walker et al.~\cite{walker2024human} define PT as consisting of regularly scheduled vehicle trips open to all paying passengers, with the capacity to carry multiple passengers whose trips may have different origins, destinations and purposes, and is ideal when the passenger regards its service as punctual and regular. As urban roadways face increasing congestion due to the growing use of cars, improving PT has become a cornerstone of traffic mitigation efforts and a critical step toward sustainable transportation~\cite{al2011composite}.

\noindent Timetabling, an essential component of PT, determines the departure and arrival times of vehicles at various stops and is crucial to providing efficient and convenient services to travelers~\cite{ceder2001efficient}. Traditionally, transit agencies develop these schedules by utilizing travel behavior surveys, such as household travel diaries and manual boarding-alighting counts. In this workflow, survey data are used to estimate average passenger loads during specific periods (e.g., morning and afternoon peaks). These estimates then form the basis for determining service frequencies and the spacing of trips within a static timetable to meet the perceived demand.

\noindent However, these current timetabling practices are often suboptimal. Because they rely on travel behavior surveys that represent only a temporal "snapshot," they frequently lack sufficient spatiotemporal coverage, are costly and time-consuming to execute, and fail to generate enough data to accurately capture dynamic demand fluctuations and PT growth~\cite{armoogum2018workshop}. Consequently, schedules derived from these methods may not reflect actual mobility patterns, leading to overcrowding during unexpected surges or inefficient service during periods of low demand.

\noindent An essential component of schedule planning is determining the frequency of bus departures, as inadequate frequency settings can lead to occupancy imbalances and service inefficiencies~\cite{hadas2012public}. To set frequencies, it is necessary to take into account the accumulated number of passengers per hour, the average travel time, the vehicle capacity given, the desired ridership and the minimum frequency allowed by time of day  \cite{ceder2016public}. These methods aim to align bus departure intervals with observed passenger loads, making sure that bus services are not underutilized or overcrowded. 

\noindent The emergence of a new era in urban research, driven by advances in big data analytics, has been highlighted by ~\citet{kandt2021smart}. These advances enable better informed decision making and provide a deeper understanding of urban systems. This shift is fueled by the rapid growth in the volume, velocity, and variety of big data, along with significant progress in data science. Innovations such as advanced data mining tools and powerful cloud computing technologies~\cite{li2015towards, li2018smart} have opened up new possibilities for analyzing passenger behavior patterns over long periods and in large urban areas ~\cite{ma2017understanding}. The availability of big data has huge potential to improve transportation planning and research. Using big data analytics and mining methods, creating an optimal schedule has become more achievable~\cite{ma2013mining}.

\noindent To capitalize on these advances and create the reliable model for predicting bus ridership, the availability of accurate and comprehensive PT datasets is essential~\cite{liang2024urban,zannat2019emerging}. Numerous databases have been developed to enhance the understanding of PT systems at various levels. Among these, the General Transit Feed Specification (GTFS)~\cite{mchugh2013pioneering} provides comprehensive schedules and route information for all PT lines planned for each day of the month. In addition, Smart Card Payment Systems for Automatic Fare Collection (AFC) employ the TAP protocol—a secure, contactless communication process that enables passengers to validate entry and exit by “tapping” a smart card or compatible device on a reader. This protocol ensures accurate fare calculation and streamlined boarding, allowing passengers to pay for their trips using smart cards at any location across the country. Moreover, laser-based Automatic Passenger Counter (APC) systems are offline devices that use laser sensors to record the number of passengers boarding and alighting at each stop. Finally, the Service Interface for Real Time Information (SIRI) dataset provides machine-readable standardized information on PT operations, including both real-time and scheduled data such as vehicle locations, estimated arrival and departure times, and service disruptions. Designed for interoperability between transit agencies, SIRI enables consistent monitoring and analysis of operational performance, supporting both passenger information systems and transportation planning. All the databases mentioned above can be integrated to provide a comprehensive understanding of the PT system at its various levels~\cite{kusakabe2014behavioural}.

\noindent In this study, we present a novel and generic methodology for long-term bus ridership prediction, capable of forecasting up to one week in advance the number of passengers remaining on board after each station along the trip, following boardings and alightings at that stop. The approach does not rely on explicit identifiers of specific lines or stops—such as route numbers, stop names, or geographic coordinates, therefore making it applicable across different PT systems. Drawing inspiration from Tobler's First Law of Geography, which states that "everything is related to everything else, but near things are more related than distant things"~\cite{miller2004tobler}. We hypothesize that predictive accuracy is improved by using localized models trained on data from spatial clusters, compared to a single global model.

\noindent Traditional approaches often impose rigid global assumptions that overlook the inherent spatial heterogeneity of urban demand. To address this, our framework integrates \textit{max-p regions} with tree-based ensembles to create data-driven boundaries reflecting real-world travel behavior. While this approach increases computational complexity by maintaining multiple sub-models, it allows for a localized capturing of demand drivers—such as the distinct behaviors of city centers versus peripheral zones—that a one-size-fits-all global regressor typically fails to resolve

\noindent To extract comprehensive information for the prediction task, data were obtained from the city’s APC dataset, which contains details on bus line, departure date, day of the week, as well as the number of passengers boarding, alighting, and the ridership of each bus at every stop along its trip. Additional datasets were incorporated, including meteorological data, which can help explain variations in PT usage across different days, along with each stop’s proximity to the city’s urban centers, which can help explain how and why different bus stops attract varying numbers of passengers. Lastly, hourly network-based representations of the PT system were constructed, allowing the extraction of network centrality measures, as well as edge weights and their statistics for ridership, distance, and related factors. These features facilitate the generalization of the model beyond the specific city context and capture the structural characteristics of the PT network within the dataset.

\noindent After collecting and analyzing data from all sources, the city was divided into spatially contiguous regions using the Max-p spatial clustering algorithm~\cite{duque2012max}. In our use case, the algorithm was applied to create polygons that group nearby stops into connected regions while ensuring that each polygon meets a minimum threshold for the sum of average passenger flows. This approach was specifically intended to form clusters of geographically proximate stops with broadly comparable ridership levels, defined as similar average total ridership across all stops within each polygon, thereby enabling localized modeling of transit patterns.

\noindent For each polygon generated by the Max-p algorithm, a separate regression model was trained using the selected input features to predict ridership at each stop immediately after departure, that is, after passenger boardings and alightings had occurre. Two main experiments were conducted to assess whether the models depended on PT–specific identifiers. The first experiment used the complete set of input features, while the second excluded all explicit identifying information—such as route ID, direction, bus stop identifier, and stop coordinates—to evaluate the models’ ability to generalize beyond the specific transit network. In parallel, global regression models were also trained on the entire dataset to serve as a benchmark and to assess whether the spatial clustering provided by the Max-p polygons improve the predictive performance. Afterwards, we applied SHAP (SHapley Additive exPlanations) on the models~\cite{nohara2019explanation} to assess the importance of the features and gain insight into the models decision-making process. 

\noindent To evaluate the effectiveness of the proposed methodology, we conducted a real-world case study using approximately 30 weeks of bus operations in the city of Be’er Sheva, consisting of more than 7.4 million stop-level records. The predictive performance of models trained separately on spatially defined polygons was compared with the same regression algorithms trained on the entire city as a single unit.

\noindent The predictive performance of the models was evaluated using a suite of error metrics, including Mean Absolute Error (MAE), Root Mean Squared Error (RMSE), Mean Percentage Absolute Error (MPAE), and their normalized variants. To ensure a granular assessment of accuracy, these metrics were first computed at the individual bus stop level by comparing predicted and observed passenger counts. This stop-specific approach provides a localized measure of model reliability across the network. These local errors were subsequently aggregated to provide a summary of overall performance, with evaluations conducted at both the hourly level and across the predefined ridership bins. The results indicated that the predictive accuracy of the global and polygon-based approaches was largely comparable, despite the differences in their underlying spatial structures.

\noindent To further assess the transferability of the methodology, we performed additional experiments without using any identifier features, namely stop ID, route ID, and stop coordinates. The results showed that the performance of the polygon clustered models remained stable even in the absence of these identifiers (see Section~\ref{sec:results}).

\begin{figure*}[!t]
    \centering
    \includegraphics[width=\textwidth, height=0.125\textheight]{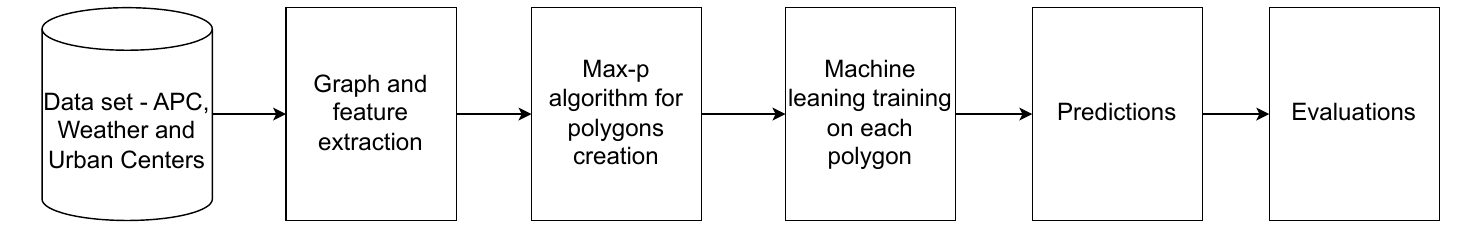}
    \caption{Methodology overview.}
    \label{fig:methodology_overview}
\end{figure*}

\noindent This study makes three primary contributions to the literature. First, it introduces a novel and generalizable framework for bus ridership prediction. The results show that the localized modeling approach achieves accuracy levels comparable to global models and performs statistically significantly better in terms of mean absolute error across multiple test sets. This finding underscores the benefit of spatially aware modeling for capturing local behavioral patterns and enhancing prediction accuracy. Second, the approach maintains robust predictive performance even when explicit identifier features such as route ID, stop ID, or stop coordinates are omitted. Third, the study analyzes the most influential features in the prediction process, providing insights into the key factors that drive model predictions.

\noindent The remainder of this paper is organized as follows: Section~\ref{sec:related-work} reviews related work on PT scheduling and approaches to ridership prediction. Section~\ref{sec:methodolog-sec} presents the proposed methodology and the generic modeling framework. Section~\ref{sec:Experiments} describes the experimental design, including the bus ridership dataset from Be’er Sheva, the pre-processing and data cleaning steps applied to ensure data quality, the setup of the training and testing splits, and the evaluation metrics used to assess model performance. Section~\ref{sec:results} reports the results, comparing the performance of spatially aware models with that of non-spatial baselines. Section~\ref{sec:discussion} discusses the implications of the findings and interpenetrate the results of the experiment, and outlines the limitation of the APC dataset. section~\ref{sec:Future Work} concludes with directions for future work. Finally, Section~\ref{sec: Conclusions} presents the main conclusions and key takeaways from this study.

\section{Related Work} 
\label{sec:related-work}
Accurate ridership prediction is vital for optimizing PT scheduling, resource allocation, and service reliability. This section provides a comprehensive review of the existing literature, organized into three main areas.

\subsection{Statistical and Machine Learning Approaches}
\label{subsec:PT prediction models} 
Classic statistic methods are based on distribution and relationship assumptions. \citet{wood2023development} predicted passenger occupancy in Pennsylvania using APC and weather data, comparing Linear Regression and Random Forest models. They evaluated performance across three levels: next-stop, origin-to-future stop, and segment-based predictions. Both models performed similarly, with minimal differences across the OD frameworks. \citet{wiecek2019framework} applied Markov chains to predict bus occupancy using APC data from a high-frequency bus line in Cracow, Poland. The model captured occupancy transitions based on previous states. The results showed that Markov chains can effectively forecast bus ridership and occupancy patterns. \citet{liu2021research} applied Auto regressive Integrated Moving Average (ARIMA) model to predict short-term passenger flow in urban rail transit systems. Comparing to Back Propagation Neural Network model, The ARIMA model significantly outperformed the BP model in both RMSE and MAE metrics. \citet{liang2019short} proposed a hybrid model that combined Kalman filters and K-Nearest Neighbors (KNN) to predict short-term passenger occupancy. The model adaptively weighted each method based on recent performance. Using data from Changchun’s Line 3 Light Rail, the approach outperformed the support vector machine (SVM) baseline, achieving lower RMSE and improved accuracy. 
    
\noindent Machine learning approaches are capable of modeling complex non linear patterns and utilizing large datasets to enhance predictive performance.  
Among them are tree - based machine learning models. For example, a popular model is XGBoost (Extreme Gradient Boosting). In contrast to deep learning architectures, which often function as black-box models, the tree-based models used are highly interpretable in the sense that predictions can be decomposed into explicit decision rules and feature contributions, allowing direct inspection of how individual variables influence model outputs. \citet{zou2022passenger} applied an XGBoost model to predict short - term bus passenger flow in Guangzhou, China, utilizing smart card data. Their model outperformed other machine learning approaches, achieving the lowest Mean Absolute Error MAE and RMSE among the compared models. \citet{barbareschi2025designing} applied XGBoost to APC data from a major Italian city, achieving prediction accuracy comparable to ARIMA and Random Forest, and outperforming deep learning models in efficiency and interpretability. They also used SHAP to explain model predictions, though deep learning methods have since gained popularity and effectiveness in bus ridership forecasting. 

\noindent Since PT demand is strongly influenced by stop location, many researchers have developed spatially aware models that incorporate geographic information into their features. These models are based on Tobler's first law (see \cite{miller2004tobler}), which state that geographically closer or more similar areas tend to exhibit similar patterns. 

\noindent \citet{wang2023bus} used Geographically Weighted Regression (GWR), which is a method that shows how the relationship between variables can change from one place to another by fitting a separate regression model at each location, to predict bus ridership at community level in the city of Beijing. The purpose of the use was to analyze how the spatial variable (such as population or employment density) may change the bus ridership in each community. Additionally, \citet{he2021adapted} applied the Adapted Geographically Weighted Lasso (Ada-GWL), which combines the spatial flexibility of GWR with the ability of Lasso to handle multi-collinearity. Tested on Shenzhen subway data, it outperformed traditional GWR with Lasso and OLS in both accuracy and stability.

\noindent Another way to capture the complex spatial characteristics of a PT system is to represent it as a network and extract meaningful topological measures from it, an approach that we also adopted in our study. For example, \citet{amilajayasinghe2014centrality} applied centrality measures such as closeness, straightness, and betweenness to enrich linear regression models for estimating boardings and alightings at PT stops. Similarly, \citet{liu2022exploring} analyzed around 700,000 smart card records from the Shanghai Metro to study factors affecting bus–metro transfer ridership. They combined land use data with network centrality measures and used an XGBoost model for prediction. Closeness centrality and bus network density emerged as the most influential factors. These studies demonstrate the predictive power of integrating network topology metrics with traditional features to model public transit ridership, providing a foundation for our own study.

\noindent \citet{kopsidas2023exploring} examined how centrality measures from metro and alternative PT networks relate to metro stop passenger flows. Using fare data and travel times from Google Maps, they modeled the metro as an unweighted graph and the alternative network (e.g., buses) as a weighted graph. Centrality features were inputted into the linear regression and XGBoost models, showing that the centrality measures of both networks significantly improved prediction accuracy.    

\subsection{Deep Learning Approaches}
\label{subsec:deep learning prediction models}
Deep learning models, a subset of machine learning, have gained increasing popularity and widespread use in recent years. A foundational example is the Multilayer Perceptron (MLP), a simple yet effective neural network architecture. \citet{farahmand2023predicting} demonstrated how the MLP model can incorporate meteorological characteristics such as temperature, precipitation, and wind speed to further enhance the predictions of bus ridership. \citet{gallo2019artificial} employed a standard feedforward Artificial Neural Network (ANN) to forecast short-term passenger flows on Naples Metro Line 1. The ANN was trained using simulation data, which was generated based on passenger counts at station turnstiles. The authors were able to estimate onboard passengers on each track section of the line as a function of turnstile data collected in previous periods. The numerical results showed that the proposed approach could forecast passenger flows on metro sections with satisfactory precision.

\noindent Some deep learning models are well-suited for short-term time series forecasting, such as predicting bus ridership minutes to an hour ahead. However, they are rarely used for longer-term forecasts due to error accumulation and the diminishing relevance of short-term patterns over time. In contrast, our study focuses on long-term predictions, looking days ahead. An example of such model is Long Short Term Memory (LSTM)(see ~\cite{gers2000learning}).  \citet{siswanto2024deep} proposes an LSTM model to predict the number of passengers for four PT bus operators and their correlation to each other. The model used historical ridership to perform the prediction task. \citet{chen2020forecasting} proposed an STL-LSTM model that decomposes the ridership data into trend, seasonal, and residual components, each modeled by separate LSTM networks. Using AFC data from Fuzhou Metro Line 1 in 2018, the approach outperformed ARIMA, SVR, and standard LSTM models in terms of RMSE. \citet{liu2019deeppf} developed a deep learning model combining LSTM and Fully Connected layers to predict metro passenger flow. Using smart card data from the Nanjing Metro, and included historical data of 103 weekdays in 2016, the model integrated spatial, weather, and operational features. It outperformed traditional models like ARIMA and FNN, achieving lower RMSE and MAE.

\noindent Bidirectional Long Short-Term Memory (BiLSTM) networks are an extension of standard LSTM models, designed to capture both forward and backward temporal dependencies in sequential data. \citet{liyanage2022ai} evaluated various neural network models for short-term bus passenger demand forecasting using smart card data from Melbourne’s PT system. Testing across 15, 30, and 60-minute intervals, they found that BiLSTM outperformed standard LSTM and other benchmark models, achieving the lowest MAE and RMSE.

\noindent Convolutional Neural Networks (CNNs) are a class of deep learning models originally developed for image recognition tasks but have since been widely applied to spatio temporal data analysis. By applying convolutional filters across spatial dimensions (e.g., grids of bus stops or metro stops), CNNs can automatically extract meaningful patterns such as flow concentrations, regional demand clusters, and localized peak periods. These spatial features can then be combined with temporal models (such as LSTM or GRU) to enhance the overall prediction of passenger flow dynamics. \citet{wu2023short} proposed a CNN-LSTM hybrid model to capture both spatial and temporal patterns in the prediction of short-term passenger flow. Using data from Foshan Bus Route No. 151, including weather, time, and passenger behavior factors, the model outperformed standalone CNN and LSTM approaches in terms of RMSE and MAE.

\noindent To test various deep learning models for short-term passenger flow forecasting, \citet{zhang2024comparative} used hourly bus data from Guangzhou, China. They performed a comparative analysis of temporal, spatial, and hybrid models, including LSTM, GRU, BiLSTM, BiGRU, CNN1D, CNN2D, CNN1D-LSTM, CNN1D-GRU, and ConvLSTM. The study concluded that bidirectional models, specifically BiLSTM and BiGRU, achieved the best overall performance in terms of accuracy.

\noindent Understanding the underlying decision processes of predictive models and identifying the factors that drive higher or lower bus ridership is extremely valuable for decision makers and transit planners. Although regression-based machine learning models offer a certain degree of interpretation, deep learning models often function as black boxes (see \citet{von2021transparency}). To overcome this challenge, \citet{monje2022deep} introduced a novel methodology using surrogate decision trees to approximate the behavior of an LSTM model trained on bus ridership data. This approach enhances the interpretation by translating the complex decision boundaries of the LSTM into human-understandable rules, thereby making the model's predictions more transparent and actionable.

\subsection{Hybrid and Combined Architectures}
\label{subsec:Advanced combo prediction models}
Many combinations of different machine learning and deep learning models were attempted to try and leverage the benefits of each model, to form new  models. For example, \citet{liu2017novel} used staked autoencoders via unsupervised learning to learn new features of the APC data. This can be used to train a DNN, or ANN, by supervised learning to predict passenger flow in train stops. \citet{chen2022prediction} divided the historical data into three categories of recent, weekly, and daily, with a PT graph that considers the similarity between the different stops. Then, using three networks constructed with bi-LSTMs, and a transformer to consider both time and spatial features altogether. Each network receives an input of one of the historical data categories and combines the results to predict the passenger ridership numbers at each stop. 

\noindent \citet{guo2019short} proposed a hybrid SVR–LSTM model for short-term abnormal metro passenger flow prediction, combining periodic patterns from SVR with recent data to extract anomalies. The fused outputs significantly outperformed ARIMA, standalone models, and fusion-KNN for 15- and 30-minute forecasts. \citet{yang2023short} utilized a multi-modal dataset from a transportation hub to predict passenger flow for three separate systems: trunk rail, intercity rail and subway. They proposed a multitask learning model based on an improved Transformer and residual network to capture complex spatio-temporal dependencies. The model demonstrated superior predictive performance, with an average prediction accuracy of 88.57\%, outperforming other baselines.
\citet{sengupta2023hybrid} demonstrate the use of a Hidden Markov Model (HMM) which is deployed to capture the stochastic nature of bus ridership. The state probabilities derived from the HMM are then used as input to an LSTM network, which captures the temporal dependencies for more accurate predictions. It was shown that in 5-minute interval prediction the mixed model results were significantly better than the models separately. 

\noindent \citet{singh2025novel} presented models for the prediction of short-term traffic flow (volume) on England's highways. Using data composing from traffic flow, speed and daily time, a combination of 1-CNN and GRU was used to capture the spatial and temporal features all together. The combined model shows better results than ARIMA and SVM, according to the RMSE metric.

\noindent \citet{zhangrj2021short} proposed a GA-TCN model that combines Temporal Convolutional Networks with a Genetic Algorithm for hyperparameter optimization. The model captures long-range temporal dependencies using causal and dilated convolutions. Trained on AIS vessel movement data from San Francisco, spanning from June 1, 2022, to December 14, 2022. It outperformed LSTM, GRU, and standard TCN models, achieving up to 27\% improvement in RMSE, MSE, and MAPE. \citet{xu2022passenger} introduced a GCN–RNN hybrid model to forecast passenger flow using 15-minute interval data from a scenic area in Beijing. The GCN captured spatial relationships among transit stops, while the RNN modeled temporal patterns. The model outperformed the baselines, improving MAE by 20\% and MSE by 40\%.

\noindent \citet{xu2022passenger} utilized a dataset of passenger flow records from a scenic area in Beijing and its surrounding transit stops, collected at 15-minute intervals. The authors proposed a hybrid GCN–RNN model to jointly capture spatial and temporal dependencies. The model's graph structure represented the scenic location and transit stops, and its output outperformed traditional models such as standalone RNN and LSTM by 20\% in MAE and 40\% in MSE.

\noindent Besides deep learning models, several approaches transform raw time-series data for traditional machine learning models. \citet{zhang2020lightgbm} introduced a hybrid model combining Variational Mode Decomposition (VMD) and LightGBM for short-term metro passenger volume forecasting. By decomposing time series data into multiple functions, each modeled by LightGBM, their approach achieved better scores in MAE and RMSE compared to traditional methods like ARIMA. In another study, \citet{sun2015novel} presented a hybrid approach that integrates wavelet analysis with LS-SVM for forecasting rail transit passenger flow. The methodology involved decomposing the original time series into low-frequency and high-frequency components, which were then modeled separately using LS-SVM. Finally, the predicted components were combined to produce the overall forecast, capturing both long and short-term patterns.

\noindent \citet{marinas2022clustering} used Salamanca PT data to group bus stops by temporal demand patterns through K-means and applied ARIMA and ANN models within each group. While prior work has also leveraged clustering to improve forecasting, a key limitation lies in the lack of spatial continuity. In contrast, our approach applies the Max-p algorithm using the Gabriel proximity graph. The Max-p framework requires an explicit spatial contiguity structure, which can be defined through several options such as rook/queen adjacency, distance bands, or proximity graphs like Delaunay, Relative Neighborhood, and Gabriel~\cite{duque2012max}. Classic work on proximity graphs highlights how these structures differ in density and inclusion relations, with Gabriel graphs offering a balanced representation: denser than the Relative Neighborhood Graph yet sparser and less redundant than the Delaunay triangulation~\cite{jaromczyk2002relative}. Importantly for geographic applications, Gabriel graphs preserve planarity, enforce strictly local connections through the empty-diameter circle criterion, and reduce unnecessary long-range edges, properties that make them particularly suitable for clustering points into contiguous, spatially coherent regions~\cite{Matula1980}.

\section{Methods}
\label{sec:methodolog-sec}
This section provides an overview of the proposed framework. The model is designed to address the specific problem of predicting \textit{bus ridership}. Specifically, given a bus, stop $X$, and time $T$, the objective is to determine the passenger load as the vehicle departs the station. By accurately forecasting the bus ridership, the model provides a critical metric to improve PT system efficiency.

\noindent The framework is organized into three stages. First, in Section~\ref{subsec:Data Collection and Preprocessing}, we describe the process of gathering and preprocessing the data. This begins with collection of the core dataset, which contains detailed information on bus rides and ridership at the stop level, followed by the integration of additional open-source datasets to enrich the feature space. Next, in Section~\ref{subsec:Data Analysis}, we outline the methods used to generate  meaningful features from the raw data and provide an overview of the approach taken to develop the prediction models. Finally, in Section~\ref{subsec:ModelEvaluation}, we present the evaluation methodology, detailing how predictive performance is assessed, and how feature importance is analyzed to determine each feature’s contribution to the model’s predictions.

\subsection{Data Collection and Preprocessing}
\label{subsec:Data Collection and Preprocessing}

The initial phase of the framework involves collecting the Automatic Passenger Count (APC) dataset for a selected public transport (PT) system. The APC dataset contains the bus line number, departure time, day of the week, and the number of passengers boarding, alighting, and remaining on board at each stop.

\noindent We represent departure time as the number of minutes since midnight, and apply sine and cosine transformations to capture its cyclic nature, making late night and early morning adjacent in meaning, and we also create categorical time-of-day bins distinguishing morning peak (06:00 to 09:00), midday off-peak (09:00 to 15:00), afternoon peak (15:00 to 18:00), evening/night off-peak (18:00 to 23:00) and night/early morning (23:00 to 06:00).

\noindent Passenger counts may be inaccurate because of limitations in the Automatic Passenger Counting (APC) system, which relies on laser sensors at bus doors. To mitigate these limitations, trips with negative ridership values and those exceeding a predefined threshold $\delta$, based on bus capacity, were excluded from the analysis.

\noindent To contextualize the passenger counts for each bus route, stop, and departure time, we collected complementary data from open sources. These include meteorological variables (temperature, rainfall, relative humidity), urban facility data (markets, educational institutions, healthcare facilities, community centers, sports centers, and high-tech employment zones), and spatial information on public transport stops (geographic coordinates, neighborhood affiliation, and socioeconomic score of the neighborhood). Together, these datasets add value by improving our understanding of ridership fluctuations across different days and hours.

\noindent We first integrate the weather dataset, where each observation corresponds to a fixed time window (e.g., hourly or 10-minute intervals). Weather conditions are temporally aligned with the scheduled departure time of each trip, ensuring consistency between environmental factors and bus trips. This is a key step, as the weather strongly influences rider decisions and demand patterns.

\noindent The spatial dataset of PT stops provides location (latitude / longitude), neighborhood, and socioeconomic score. The latter serves as a proxy for the demographic profile and travel behavior of each neighborhood—for example, areas with lower socioeconomic scores may show a higher reliance on public transport due to reduced car ownership rates. This information is integrated by linking each APC observation to the corresponding stop attributes.

\noindent Finally, the urban facilities dataset captures the main travel destinations, such as schools, health services, markets, gyms, and community centers. These facilities highlight the functional role of different areas and help explain demand peaks related to everyday activities.

\noindent To merge the urban facilities dataset with the APC dataset, it is necessary to determine, for each stop–facility pair, whether the distance between them is small enough for the stop to reasonably serve as a source or destination to reach the facility. Therefore, for each observation in the APC dataset, representing a specific bus stop~\(s\) and a radius $r$, we define a spatial buffer centered at~\(s\), where all urban facilities located within this buffer are identified, counted, and categorized by type (e.g., medical centers, educational institutions, sports facilities, markets, etc.). $r$ can be adjusted according to the urban context or the research objectives. This spatial information allowed the creation of new features that help assess how the density and variety of nearby facilities can influence the number of bus passengers.

\subsection{Data Analysis}
\label{subsec:Data Analysis}
The data analysis comprises three primary stages. First, the PT network is modeled as hourly directed graphs to capture the temporal dynamics of bus operations.  Second, a comprehensive set of features is extracted from these graphs, including centrality measures, edge-level statistics, and descriptive indicators that characterize both stops and routes. These features are integrated with temporal, meteorological, and spatial attributes to construct the analytical dataset. Lastly, the city is divided into spatially contiguous regions using the Max-p algorithm, which groups proximate stops into polygons that are homogeneous in terms of ridership patterns. This multi-stage approach facilitates simultaneous analysis of structural, temporal, and spatial aspects of ridership.

\noindent The rest of this section is organized as follows: Subsection~\ref{subsec:graph_creation} illustrates the graph-creation procedure, describing how hourly network snapshots are constructed and enriched with passenger-related and structural attributes. Subsection~\ref{subsec:feature_extraction} outlines the feature extraction process, where graph-based, temporal, meteorological, and spatial variables are integrated into the analytical dataset. Subsection~\ref{subsec:polygon_creation} presents the polygon creation steps and the modeling frameworks, in which the Max-p algorithm partitions the city into spatially contiguous regions with homogeneous ridership patterns, and predictions are generated using both polygon-based and global models.

\subsubsection{Graphs creation}
\label{subsec:graph_creation}
 
To obtain deeper insights into the characteristics of the PT system from the raw datasets, we construct complex networks that represent the PT at different times of the day. Specifically, hourly directed graphs of the city's bus network are generated. This temporal resolution is chosen because stations and routes can vary significantly in both characteristics and importance depending on the time of day. For example, routes serving schools and industrial areas are more critical in the morning, whereas connections to markets and restaurants become more significant in the afternoon and evening.  

\noindent Let \(T = [t, t+\Delta t)\) denote the time interval. For each such interval, we construct a directional graph \(G_T = (V, E_T)\). Each such graph reflects the observed structure and activity of the PT network during \(T\), rather than a forecast of future network conditions. These graphs serve as the basis for generating explanatory features that provide the model with structural and temporal context. In this framework, the variable \texttt{PassengersContinue} is used as a proxy for bus ridership, since it represents the number of passengers remaining onboard between consecutive stops. For example, to predict ridership for a ride departing at 14:30 on line~1, we use the network snapshot corresponding to \(T = [14{:}00,15{:}00)\) and extract centrality and edge-based features for the relevant stops. Formally, for each reference time \(T\), we define the snapshot \(G_T = (V, E_T)\) by:
\begin{itemize}[leftmargin=*]
    \item \( V \): the set of vertices (bus stops), time-invariant.
    \item \( E_T \subseteq V \times V \): directed edges, where \((u,v)\in E_T\) if and only if there exists at least one trip during the interval \(T\) with \(u\) and \(v\) as consecutive stops.
\end{itemize}

\noindent Each edge \((u,v)\) carries the following weights:
\begin{enumerate}[leftmargin=*]
\item \texttt{number of routes}: $r_T(u,v)$ is the number of \newline distinct routes active during \(T\) that include \(u\) and \(v\) as consecutive stops.
\item \texttt{distance}: $d(u,v)$ is the geodesic \newline distance (in meters) between stops \(u\) and \(v\), time-invariant.
\item \texttt{PassengersUpSum\_sum}: $B^{\mathrm{sum}}_T(u,v)$ is the total \newline
boardings on segment \(u \to v\) during \(T\).
\item \texttt{PassengersUpSum\_avg}: $B^{\mathrm{avg}}_T(u,v)$ is the mean \newline
boardings per trip on \(u \to v\) during \(T\).
\item \texttt{PassengersUpSum\_std}: $B^{\mathrm{std}}_T(u,v)$ is the standard deviation of boardings per trip on \(u \to v\).
\item \texttt{PassengersDownSum\_sum}: $A^{\mathrm{sum}}_T(u,v)$ is the total alightings on \(u \to v\) during \(T\).
\item \texttt{PassengersDownSum\_avg}: $A^{\mathrm{avg}}_T(u,v)$ is the mean alightings per trip on \(u \to v\) during \(T\).
\item \texttt{PassengersDownSum\_std}: $A^{\mathrm{std}}_T(u,v)$ is the standard deviation of alightings per trip on \(u \to v\).
\item \texttt{PassengersContinue\_sum}: $C^{\mathrm{sum}}_T(u,v)$ is the total through-passengers (i.e., bus ridership) on \(u \to v\) during \(T\).
\item \texttt{PassengersContinue\_avg}: $C^{\mathrm{avg}}_T(u,v)$ is the mean through-passengers per trip on \(u \to v\).
\item \texttt{PassengersContinue\_std}: $C^{\mathrm{std}}_T(u,v)$ is the standard deviation of through-passengers per trip on \(u \to v\).
\item \texttt{total\_segment\_frequency}: $f_T(u,v)$ is the number of observed trips traversing \(u \to v\) during \(T\).
\item \texttt{total\_segment\_frequency\_bin}: This is a categorical bin of $f_T(u,v)$ obtained by uniform binning into 5 categories (e.g., very low, low, medium, high, very high).
\end{enumerate}

\noindent These graphs provide a deeper understanding of how the PT system behaves during different times of the day, and from that we can extract information to feed predictive models in order to forecast the bus ridership. For the remainder of this section, we use as a guiding example a weekday trip on line~1 departing at 14{:}30 arriving at stop~\(s\). Our goal is to predict the ridership at station~\(s\) immediately after the bus departs, i.e., following passenger boardings and alightings at that stop.

\subsubsection{Feature Extraction}
\label{subsec:feature_extraction}

\noindent From the PT network graphs, we compute standard centrality measures, such as degree, closeness, betweenness, and eigenvector centrality, to quantify each stop’s structural importance~\cite{zhang2017degree, bonacich2007some}. Two additional network-level measures were also included. Density~\cite{anderson1999interaction} represents the proportion of possible edges that are actually present, providing an indicator of overall connectivity (higher values correspond to more links between stops). The second measure, the average clustering coefficient~\cite{watts1998collective}, captures the extent to which the neighbors of a stop are also connected to one another. This reflects the local cohesion of the network and indicates the potential for transfers, with higher values signifying more tightly connected neighborhoods. 

\noindent Descriptive statistics, including sum, average, median, minimum, and maximum, were computed at both the network and route levels to supplement centrality and global network metrics. At the network level, passenger counts (onboarding, alighting, and ridership), inter-stop distances, and trip frequencies were summarized across all segments to capture system-wide trends. 

\noindent At the route level, these statistics were calculated separately for each bus line and for the segments between its stops to serve as the primary historical features for the machine learning models. These route and segment-level (edge) statistics are operationalized as follows:

\begin{itemize}[leftmargin=*]
    \item \textbf{Route-level Features:} These represent the historical behavior of a specific bus line across the entire network. For example,  \texttt{Route\allowbreak\_continuing\allowbreak\_passengers\allowbreak\_avg} is the historical average of passengers remaining on board after serving a stop on that specific route. This allows the model to identify high-demand lines regardless of the specific stop location.
    \item \textbf{Edge-level (Segment) Features:} These represent the historical load on the specific physical segment (edge) between two stops.\newline For example, \texttt{continuing\allowbreak\_passengers\allowbreak\_on\allowbreak\_edge\allowbreak\_avg} describes the typical occupancy level seen on a specific stretch of road, which can be shared by multiple routes.
    \item \textbf{Onboarding/Alighting Averages:} Features such as \newline\texttt{Boarding\allowbreak\_passengers\allowbreak\_on\allowbreak\_edge\allowbreak\_avg} capture the typical passenger exchange at a specific location, signaling likely changes in total occupancy for the subsequent trip segment.
\end{itemize}

\noindent To illustrate the distinction between these features, consider a trip on \textbf{Line 1} traveling between \textbf{Stop A} and \textbf{Stop B}:
\begin{enumerate}[leftmargin=*]
    \item \textbf{Route Perspective:} The feature \texttt{Route\_continuing\allowbreak\_passengers\allowbreak\_avg} reflects Line 1's general popularity. If Line 1 is a major city-wide trunk line, this feature informs the model that the vehicle is likely to carry a high volume of passengers throughout its entire journey.
    \item \textbf{Edge Perspective:}  \texttt{Continuing\_passengers\allowbreak\_on\allowbreak\_edge\_avg} focuses on the physical road segment between Stop A and Stop B. If this segment is a high-traffic corridor used by multiple different bus lines (e.g., Line 1, Line 5, and Line 10), this feature informs the model that any bus on this specific stretch of road historically carries a heavy load, regardless of its specific route number.
    \item \textbf{Local Exchange:} If Stop A is located near a residential hub, a high \texttt{Boarding\_passengers\_on\_edge\_avg} value at that specific stop acts as a precursor, allowing the model to anticipate an immediate spike in \newline \texttt{PassengersContinue} for the segment leading to Stop B.
\end{enumerate}

\noindent By distinguishing between route-wide patterns and segment-specific realities, the framework can effectively differentiate between the inherent popularity of a bus line and the localized demand of the urban geography it traverses. At this stage, the complete dataset has been assembled and is ready to serve as input to construct the predictive models. 
\begin{enumerate}[leftmargin=*]
    \item Temporal and PT attributes such as bus route and departure time;
    \item Meteorological conditions at the time of the ride;
    \item Stop-level spatial attributes, including proximity to urban facilities (e.g., counts of medical, educational, and commercial institutes within a defined radius); and
    \item Graph-derived features that characterize the structural role of the route and stop within the PT network.
\end{enumerate}

\noindent To clarify the modeling framework, each observation in the dataset represents a specific bus trip at a given stop $s$ and time $t$. The inputs and outputs of the machine learning models are defined as follows:

\begin{itemize}[leftmargin=*]
    \item \textbf{Model Inputs:} A feature vector comprising the bus line, departure time, temporally aligned weather conditions, and stop-level spatial attributes, including proximity to urban facilities (e.g., counts of medical, educational, and commercial institutes within a defined radius). Additionally, it includes stop-level graph characteristics and features describing the surrounding urban environment relevant to that specific time $t$.
    \item \textbf{Model Output:} The predicted number of passengers on board the vehicle immediately after departing stop $s$. This target variable accounts for the net passenger exchange (onboard and offboard) at that location.
\end{itemize}

\noindent Collectively, these features offer a comprehensive representation of the operational and spatial context, serving as the basis for predicting bus occupancy at both the line and stop levels. This two-level approach integrates system-wide and route-specific insights to complement the structural measures derived from network graphs. The graph features were subsequently merged with the ride-level data by aligning each ride with its corresponding hourly graph and associating each segment and stop with the appropriate directed edge in the network. For instance, a bus departing at 14:30 on line 1 would be matched with the graph snapshot for T = [14:00, 15:00), and its statistics for the edge (u, v), where u and v are consecutive stops on the route, would be incorporated into the ride-level data.
Collectively, we created and utilized 127 features, which offer a comprehensive representation of the operational and spatial context of each ride and serve as input for predicting bus ridership (PassengersContinue) on the bus line and stop levels.

\noindent The input of the model in the running example is illustrated as follows. The objective is to predict ridership at stop s for a weekday departure of line 1 at 14:30. Stop~\(s\) is situated near several schools and a commercial area. Accordingly, two land-use features quantify the number of schools and commercial facilities within a specified proximity of~\(s\). To provide contextual information for this prediction, the ride is associated with the hourly graph for the interval T = [14:00, 15:00). From this graph, centrality measures for~\(s\) and edge-level statistics for the directed segment (u, s), where u is the preceding stop, are extracted. Meteorological data such as temperature and rainfall are linked to the same departure time, and stop-level attributes quantify the educational and commercial facilities within the defined proximity buffer. These features are combined to provide the model with a detailed operational and spatial profile of the ride, supporting the prediction of bus ridership at stop ~\(s\).

\subsubsection{Polygon and Global Modeling Frameworks}
\label{subsec:polygon_creation}
Two alternative frameworks are presented for bus ridership prediction. The global framework involves training a single forecasting model on the entire citywide dataset, which captures general trends but does not adapt to localized conditions. In contrast, the polygon-based framework delineates distinct, contiguous urban regions and trains a separate model for each region. These parallel procedures facilitate direct comparison between global and regional learning approaches. 

\noindent Within the polygon-based framework, this study employs the Max-p regions approach to delineate spatial clusters. This choice is made to avoid the restrictive structural assumptions inherent in hierarchical models, which typically require the \textit{a priori} identification of central stops or hubs based on specific centrality metrics. Such definitions are often network-specific and do not generalize well across different urban morphologies. By contrast, the Max-p approach is entirely data-driven, forming regions based on spatial contiguity and statistical viability. This prevents model bias associated with center-periphery assumptions and better accounts for the multiple centers of activity common in modern cities, where high-demand areas may coexist or shift depending on the time of day. Consequently, this method ensures that secondary areas are not overly smoothed by the dominance of a single hub, allowing for more granular and localized ridership predictions.

\noindent Furthermore, while the model could theoretically be implemented at the level of individual bus stops, the Max-p regionalization approach is preferred for its statistical robustness and computational efficiency. Predicting ridership for every discrete stop often introduces significant stochastic noise, as many stops suffer from sparse data or highly irregular usage patterns. By aggregating stops into statistically viable regions, the Max-p approach ensures a minimum sample size for each model, thereby enhancing stability and predictive reliability.

\noindent Additionally, this regional approach effectively captures spatial dependencies. Demand at adjacent stops is rarely independent; instead, it is influenced by shared local characteristics such as land use and network topology. Max-p enforces spatial contiguity, allowing the models to learn these shared patterns rather than treating each stop as an isolated entity. Finally, from a computational perspective, regionalization significantly improves scalability. Rather than training $N$ models for every stop in the network, the framework reduces the requirement to $P$ regional models (where $P \ll N$), making the approach feasible for large-scale urban deployments.

\noindent To facilitate this comparison, we utilize the dataset described in Section~\ref{subsec:feature_extraction}. 
The dataset contains adequate information to support the global modeling framework. However, implementing the polygon-based framework requires partitioning the city into spatially contiguous regions. Generating spatially coherent regions is essential for the polygon-wise modeling approach.
The Max-p clustering algorithm requires a threshold parameter $\tau$ that determines the minimum ridership for each region. Higher threshold values result in larger regions that group more stops together, while lower thresholds produce smaller, more detailed regions. There is no universally accepted value for $\tau$.

\noindent Therefore, $\tau$ is defined as a function of the average stopping level of traffic, ranked by a factor $k$. This formulation ensures that the threshold remains data-driven and adjustable. The Max-p algorithm was then applied on a fixed Gabriel proximity graph (see Section~\ref{sec:related-work} for each candidate threshold, defined as a function of average stop-level ridership scaled by a factor $k$ (denoted $\tau(k)$). For each resulting partition, we computed the Calinski–Harabasz (CH) index~\cite{calinski1974dendrite}, which balances between-region separation and within-region cohesion, and selected the partition that maximizes CH. This internal validity criterion is standard in the clustering of transit data of ridership~\cite{huang2019interactions, huo2022using} 

\noindent Several regression models are applied in both frameworks. Each regression model was applied in two ways: in the global framework, models were trained on the entire city, while in the polygon framework, separate models were trained for each polygon. The regression models used in the study include Linear Regression and tree-based models: LightGBM~\cite{ke2017lightgbm}, Random Forest~\cite{breiman2001random}, XGBoost~\cite{chen2016xgboost}, and CatBoost~\cite{prokhorenkova2018catboost}, are employed to predict bus ridership, using the input features described in Section~\ref{subsec:Data Analysis}.

\noindent We demonstrate how both the global and polygon-based frameworks are being applied using our running example: predicting the ridership of a weekday \textit{line~1} departure at 14{:}30, specifically at stop~\(s\).
In the global framework, a single citywide model trained on all observations in the train set, and then produces the ridership estimate for stop~\(s\).
In the polygon-based framework, after polygon creation, stop~\(s\) is assigned to polygon~\(P_j\); during prediction, the ride is matched to \(P_j\), and the corresponding region-specific model generates a ridership estimate for stop~\(s\). Predictions from all polygons \(\{P_1, P_2, \ldots, P_k\}\) are subsequently assembled into a unified citywide prediction framework.

\subsection{Model Evaluation}
\label{subsec:ModelEvaluation}



\noindent To evaluate the accuracy of both frameworks, we adopt an \emph{iterative rolling-origin} approach based on consecutive periods. Let $D$ denote the final day of a given period, and let $H_1, H_2 \in \mathbb{N}$ represent the lengths (in days) of two evaluation windows. We define:

\begin{itemize}[leftmargin=*]
    \item the \emph{first evaluation window} ${W}_{1} = [D-(H_1+H_2)+1,\; D-H_2]$,  
    \item the \emph{second evaluation window} ${W}_{2} = [D-H_2+1,\; D]$.  
\end{itemize}

\noindent The ${W}_{1}/{W}_{2}$ split is designed to simulate a real-world deployment scenario, where a model trained on historical data (${W}_{1}$) must forecast ridership for the immediately subsequent period (${W}_{2}$). By adopting this iterative rolling-origin approach, we achieve three primary objectives:

\begin{enumerate}[leftmargin=*]
    \item \textbf{Assessment of Robustness:} Public transport demand is subject to significant fluctuations driven by holidays, special events, or weather changes. Multiple train-test pairs allow for a systematic evaluation of how sensitive the model is to these external variations.
    \item \textbf{Identification of Temporal Effects:} Comparing performance across distinct windows helps identify specific periods where the model excels or struggles, highlighting the impact of unusual demand patterns (e.g., religious holidays or seasonal shifts).
    \item \textbf{Enhanced Generalization:} Rather than relying on a single temporal snapshot, testing across consecutive windows ensures that the reported accuracy represents sustained performance and long-term generalizability.
\end{enumerate}

\noindent To maintain this realistic forecasting setting, the models are repeatedly retrained as new data becomes available. Within each iteration, the training data undergoes a cleaning process to resolve Automated Passenger Counter (APC) discrepancies, specifically addressing imbalances between total boardings and alightings over a single ride. This ensures the model learns from high-fidelity inputs before being evaluated on the subsequent evaluation window. The procedure is as follows:  
\begin{enumerate}[leftmargin=*]
    \item \textbf{Step 1: Data partitioning and cleaning.}  
    Split the data into a training set (all observations up to $D-(H_1+H_2)$) and the first evaluation set ${W}_{1}$.  
    For each ride in the training data, compute the total number of boardings ($TB$) and alightings ($TA$) by summing counts across all stops along the route.  
    Discrepancies were quantified using two complementary measures:
    \begin{equation*}
    \text{Percentage Difference} =
    \frac{|TB - TA|}{\max(TB, TA)} \times 100
    \end{equation*}
    
    \begin{equation*}
        \text{Absolute Difference} =
        |TB - TA|
    \end{equation*}

 The removal of outliers is performed using a dual criterion: a ride was excluded only if both the percentage difference and the absolute difference exceeded their respective thresholds. Thresholds are determined using the interquartile range (IQR) method, based solely on the training data. Specifically, for each measure the first quartile ($Q_1$) and third quartile ($Q_3$) were compute, and the IQR was defined as
    \[
        \text{IQR} = Q_3 - Q_1.
    \]
    Values outside the range
    \[
        [\,Q_1 - 1.5 \times \text{IQR},\; Q_3 + 1.5 \times \text{IQR}\,]
    \]
    are flagged as extreme. This ensured that (i) rides with very small passenger counts were not unfairly excluded for minor proportional mismatches, and (ii) rides with genuinely large absolute mismatches were detected.  
    The cleaned training set is then used to train the model, and performance is evaluated on ${W}_{1}$, after applying the same thresholds to filter the evaluation window (without re-estimating them).+

    \item \textbf{Step 2: Iteration and retraining.}  
    Extend the training set to include ${W}_{1}$ (i.e., all data up to $D-H_2$).  
    Reapply the same IQR-based filtering procedure to this enlarged training set (with thresholds computed only from the training portion), retrain the model, and evaluate performance on ${W}_{2}$.  
\end{enumerate}

\noindent This iterative process provided a consistent and data-driven framework for model evaluation. By applying the dual discrepancy criteria before training in each step, extreme or unrepresentative rides were removed from the training data while ensuring that test sets remained unaffected by filtering, thereby preventing information leakage.

\noindent For example, (January; $H_1=H_2=7$).
Consider January 1--31 (\(D=\) Jan~31). Then
\[
{W}_1 = \text{Jan 18--24}, \qquad 
{W}_2 = \text{Jan 25--31}.
\]

\noindent The approach proceeds as follows:
\begin{enumerate}[leftmargin=*]
  \item \textbf{Step 1:}  Training data = Jan 1--17.  
  Compute $TB$ and $TA$ per ride, derive Absolute and Percentage Differences, and determine IQR-based thresholds \emph{from Jan 1--17 only}.  
  Apply these thresholds to filter both the training set and the evaluation window ${W}_1$.  
  Train the model on the cleaned training set and evaluate on the filtered ${W}_1$.
  
  \item \textbf{Step 2:}  Training data = Jan 1--24 (i.e., Step~1 train plus ${W}_1$).  
  Recompute thresholds from this enlarged set, apply them to filter both the training data and ${W}_2$, retrain the model, and evaluate on the filtered ${W}_2$.
\end{enumerate}

\noindent This scheme iteratively evaluates the models while continuously incorporating the most recent observations, closely reflecting the operational deployment. Both the global and polygon-based frameworks are applied to each train–test pair for consistent comparison.  

\noindent Model performance is evaluated using five standard error metrics: Mean Absolute Error (MAE), Root Mean Squared Error (RMSE), Mean Absolute Percentage Error (MAPE), percent RMSE (\%RMSE), and symmetric MAPE (sMAPE). The application of RMSE and MAPE is well-established in the literature on transit ridership prediction; for example, both are used in the prediction of bus passenger flow by \citet{zou2022passenger}, while studies of metro ridership have similarly adopted these as principal criteria \cite{wang2018forecasting}. 

\noindent The inclusion of sMAPE is particularly critical in this study, as stop-level ridership frequently contains zero or near-zero values. Under such conditions, standard MAPE can produce infinite or exaggerated errors, whereas sMAPE provides more stable and interpretable results by normalizing against the mean of predicted and observed values.

\noindent Although these metrics often exhibit a high degree of correlation, they are reported concurrently to provide distinct perspectives on model performance that a single measure cannot capture. This multi-metric approach is justified by three key considerations:

\begin{itemize}[leftmargin=*]
    \item \textbf{Error Sensitivity:} Different metrics emphasize distinct aspects of the error distribution. For instance, RMSE is more sensitive to outliers than MAE, allowing for an assessment of whether model inaccuracies are consistent or driven by rare, large-scale deviations.
    \item \textbf{Absolute vs. Relative Interpretation:} Absolute metrics (MAE, RMSE) quantify raw passenger count errors, which is vital for capacity planning. Conversely, relative metrics (\%RMSE, sMAPE) normalize errors against ridership volume. This is essential for evaluating performance across diverse stop types, where an error of 10 passengers may be negligible at a high-volume hub but critically significant at a low-volume stop.
    \item \textbf{Operational Relevance:} From a transit management perspective, both the magnitude and the proportionality of errors are relevant. A comprehensive evaluation ensures model reliability across varying demand intensities, from off-peak periods to rush-hour surges.
\end{itemize}

\noindent These metrics are applied consistently across all experiments, with results reported at the aggregate level for each model across all test sets. In addition to this evaluation, we examine the ability of the models to generalize across different PT systems. To assess transferability, we consider two experimental settings. The first setting uses the complete feature set, which includes system-specific identifiers such as bus line numbers, stop identifiers, and geographic coordinates. The second setting excludes these identifiers, requiring the model to rely solely on generalizable patterns.

\noindent To evaluate whether the differences in predictive performance between polygon-wise and global models were statistically significant, paired statistical tests were conducted across all experimental runs. Since each model was evaluated on identical test sets, the resulting MAE values form matched pairs suitable for dependent-sample analysis. Accordingly, the non-parametric Wilcoxon signed-rank test was applied to assess whether the observed performance differences were statistically meaningful~\cite{demvsar2006statistical}. Two comparisons were examined: polygon vs.\ global models with ID features, and polygon vs.\ global models without ID features.

\noindent To add to evaluating overall accuracy and generalization, a more detailed analysis of the models was conducted. Model performance was initially compared across all candidates using mean absolute error (MAE) as the primary metric, as MAE offers a direct measure of average prediction error. The analysis then proceeded in two directions: first, performance was stratified by temporal conditions (hour of departure) and, second, by rideship levels. To provide a clearer framework for the latter, ridership was classified into discrete 'levels' rather than treated as a continuous variable. Specifically, passenger numbers were grouped into predefined bins (e.g., 0–10, 11–20, etc.), effectively representing low, medium, and high demand states. This stratification ensures that performance fluctuations are analyzed in the context of varying load intensities. This stratification was applied to both the global and polygon-wise models to determine which framework more effectively captures temporal and demand-related patterns. Additionally, model performance was evaluated across individual months and using an iterative weekly training approach to assess potential temporal differences and to determine whether expanding the training data with observations from preceding weeks improves forecasting accuracy. Second, SHAP~\cite{nohara2019explanation} was applied exclusively to the single best-performing model to quantify the contribution of individual features to its predictions. Figure~\ref{fig:workflow} presents the overall workflow of the proposed methodology.

\section{Experiments}
\label{sec:Experiments}
This section outlines the experimental setup, which is on the Dan Be’er Sheva APC dataset, the GOV.il datasets, and the open-7 Be’er Sheva open-source datasets. We first present the size and structure of the raw data, followed by the data cleaning procedures and threshold settings applied to improve data quality. Finally, we present how the data were partitioned using the Max-p algorithm in our experiments, with thresholds derived from average ridership per stop and the optimal choice guided by the Calinski–Harabasz (CH) index.
\begin{itemize}[leftmargin=*]

\item \textbf{APC Dataset.} \phantomsection\label{Experiments:apc}
The Israeli Innovation Authority provided a primary dataset comprising 7,410,003 stop-level entries from the Dan Be'er Sheva bus network collected between November 2022 and May 2023. Each record details the route, direction, actual and scheduled departure times, weekday, stop ID, neighborhood, stop sequence, boarding and alighting counts, and total ridership.

\item \textbf{Meteorological Dataset.} \phantomsection\label{Experiments:weather}
Sourced from the GOV.il website, this dataset offers 10-minute interval data on rainfall (mm), temperature (°C), and relative humidity from November 2022 to May 2023. Each APC record is matched to the closest meteorological measurement.\footnote{Valid as of 2024, \href{https://ims.gov.il/he}{\textcolor{blue}{Meteorological Dataset}}}

\item \textbf{Bus Stops Dataset.} \phantomsection\label{Experiments:busstops}
From the GOV.il open data portal, this dataset provides metadata on 757 bus stops in Be'er Sheva, including stop ID, name, location (latitude and longitude), and the neighborhood in which each stop is located, together with the neighborhood's social score.\footnote{Bus stops dataset valid as of 2024, social score available as of 2021 \href{https://data.gov.il/dataset/bus_stops}{\textcolor{blue}{Bus stops}}, \href{https://data.gov.il/dataset/stat_n-hoods_table-br7}{\textcolor{blue}{Neighborhood ID}}, and \href{https://www.cbs.gov.il/he/publications/Pages/2025/characterization-geographic-units-and-classification-socio-economic-2021.aspx}{\textcolor{blue}{CBS Socio-economic Classification}}}

\item \textbf{Urban Facilities Dataset.} \phantomsection\label{Experiments:facilities}
The dataset comprises the locations of 38 major facility types and areas in Be’er Sheva, including markets, universities, hospitals, the old city, and the high-tech zone, sourced from the Israeli government’s open data platform. Additional datasets were incorporated from open-7 Be’er Sheva open-source datasets, to represent smaller institutions such as gyms, small sports facilities, clinics, community centers, playgrounds, and schools. The integration of these sources enabled measurement of the built environment surrounding each stop.\footnote{Facilities dataset (valid as of 2024). Available at:
\href{https://data.gov.il/dataset/playground-datagov-br7}{Playgrounds},
\href{https://data.gov.il/dataset/businesscenters-br7}{Business Centers},
\href{https://data.gov.il/dataset/sport-br7}{Sports Facilities},
\href{https://data.gov.il/dataset/community-centers-br7}{Community Centers},
\href{https://data.gov.il/dataset/health-br7}{Health Facilities},
\href{https://data.gov.il/dataset/elderly-social-clubs-br7}{Elderly Clubs},
\href{https://data.gov.il/dataset/education-br7}{Education}, and
\href{https://data.gov.il/dataset/daycares-br7}{Day Cares}.}

\end{itemize}

\noindent The APC dateset was pre-processed to address inconsistencies in passenger counts at the stop level, as outlined in Section~\ref{sec:methodolog-sec}. The most prominent issue involved negative passenger counts, attributable to technical malfunctions. Of the 220,788 rides recorded, 103,641 contained at least one stop with a negative value. In addition, to account for the maximum capacity of inner-city buses, we introduced a ride-level threshold of $\delta = 50$. Any ride in which the number of boarding, alighting, or continuing passengers at a single stop exceeded $\delta$ was excluded. This procedure resulted in the removal of 647 rides.

\noindent Following data cleaning, we merged the urban facilities dataset with the APC data as described in section ~\ref{sec:methodolog-sec}. For this experiment, the spatial buffer for each bus stop $s$ was defined as $r = 200 \text{meter}$. Within this buffer, all urban facilities were identified, counted, and categorized by type, including medical centers, educational institutions, markets, and recreational areas. This radius was selected to represent a reasonable walking distance in an urban context, consistent with the widely adopted 400 meters service area for bus stops, which implies an expected average walk of roughly half that distance (around 200 meters)~\cite{el2014new, daniels2013explaining}. Table~\ref{tab:urban_facilities} provides a list of the facility types used in the analysis, along with a brief explanation of each category.

\noindent To structure the experiments, the dataset was partitioned into training and testing subsets using the iterative procedure described in Section~\ref{sec:methodolog-sec}. This was performed separately for each month, with $D$ denoting the last day of the month. We set $H_1 = H_2 = 7$, corresponding to week-long evaluation windows, since the framework aims to predict ridership one week ahead in a long-term setting. As the dataset spans seven months (November 2022--May 2023), the procedure yielded 14 distinct train–test pairs. For each iteration, outlier removal was carried out by deriving thresholds from box-plot statistics (IQR) on the training set and applying them unchanged to the corresponding test set, ensuring consistent filtering across both subsets (see Section~\ref{subsec:Data Collection and Preprocessing}).

\noindent For each train–test iteration, we assessed the optimal threshold $\tau$ for the Max-p clustering algorithm. The threshold was defined as the total sum of average ridership across all stops, divided by a scaling factor $k$. Formally,  
\[
\tau(k) = \frac{\sum_{i=1}^{N} \overline{R}_i}{k},
\]  
where $\overline{R}_i$ denotes the average ridership at stop $i$ within the training set, and $N$ is the total number of stops. We evaluated scaling factors $k \in \{5, 10, 15, 20, 25, 30, 35, 40, 45, 50\}$, which corresponds to partitions ranging from very coarse ($\sim$150 stops per region) to finer ($\sim$15 stops per region), given the 757 stops in the dataset. This range balances computational feasibility with meaningful spatial granularity. As described in Section~\ref{sec:methodolog-sec}, the Calinski–Harabasz (CH) index was used to select the final partition.

\noindent Statistical significance tests, as detailed in the Methods section, were conducted on all 14 matched test sets by month and week to determine whether the observed performance differences between polygon-wise and global models were statistically significant. Based on the evaluation results, the model with the highest average performance in all test sets was selected in terms of MAE. SHAP was then applied to identify the features that most strongly influenced its predictions. The analysis focused on forecasts for the last week of January and the last week of May, using models trained on data from the same month up to (but excluding) the respective week, and evaluated both with and without identifying features. These two periods were selected to examine whether seasonal and weather-related conditions affect the relative importance of features.

\section{Results}
\label{sec:results}
The analysis utilized the APC dataset. The dataset contains \newline 7,410,003 stop-level records from the Dan Be’er Sheva network collected between November 2022 and May 2023. Tables~\ref{tab:features_base},~\ref{tab:features_centrality},~\ref{tab:features_edge},~\ref{tab:features_weights} and~\ref{tab:features_route} summarize the complete set of features employed in the experiments. These include base attributes of stops and routes, centrality measures derived from the transit network, edge-level and route-level characteristics, as well as features related to passenger flows and service frequency. 

\noindent After the initial data cleaning described in Section~\ref{subsec:Data Collection and Preprocessing}, 3,826,398 entries remained. For each train–test iteration, outliers were removed using the IQR procedure, and the Max-p algorithm was applied to generate spatially contiguous regions. Table~\ref{tab:summary_month_week} summarizes the number of training and test samples per iteration, together with the resulting regional partitions. The number of regions directly determines how many models are trained in the polygon framework, since a separate model is fitted for each region in every iteration.

\begin{table}[t]
\centering
\small
\caption{Summary of samples per month and week.}
\label{tab:summary_month_week}
\resizebox{0.99\columnwidth}{!}{%
\begin{tabular}{|c|c|c|c|c|}
\hline
\textbf{Month} & \textbf{Last Week (0/1)} & \textbf{Train Samples} & \textbf{Test Samples} & \textbf{Num Polygons}\\\hline 1 & 1 & 404909 & 118461 & 4 \\\hline 1 & 0 & 261862 & 117871 & 7 \\\hline 2 & 1 & 473899 & 110580 & 7 \\\hline 2 & 0 & 331242 & 118442 & 4 \\\hline 3 & 1 & 435285 & 132171 & 4 \\\hline 3 & 0 & 319286 & 115158 & 8 \\\hline 4 & 1 & 369541 & 94710 & 7 \\\hline 4 & 0 & 258778 & 86529 & 3 \\\hline 5 & 1 & 567020 & 125448 & 7 \\\hline 5 & 0 & 473562 & 93458 & 7 \\\hline 11 & 1 & 378081 & 110559 & 4 \\\hline 11 & 0 & 275029 & 79722 & 4 \\\hline 12 & 1 & 290750 & 81310 & 4 \\\hline 12 & 0 & 266213 & 24537 & 4 \\\hline 
\end{tabular}
}
\end{table}

\begin{figure}[ht]
     \centering
     \includegraphics[width=\columnwidth]{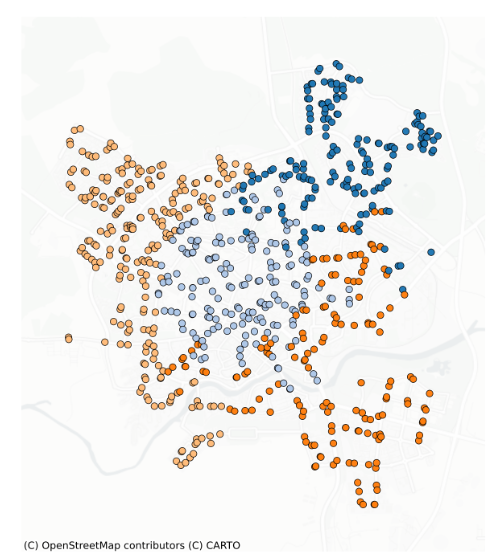}
     \caption{Max-p regions for March (training on all days except the last 7). 
  Bus stops are grouped into spatially contiguous regions, shown in different colors, 
  as determined by the optimal threshold $\tau$ selected using the Calinski–Harabasz (CH) index.}
     \label{fig:max_p_march}
\end{figure}

\noindent As an illustrative example, Figure~\ref{fig:max_p_march} shows the Max-p partitioning of the 757 bus stops for March (excluding the last 7 days of the month), which produced four large regions. The resulting clusters display four distinct spatial patterns: northern and eastern stops form one contiguous block, southern stops form another, and the city center with adjacent neighborhoods are grouped together. This example demonstrates that the selected threshold yields interpretable regions consistent with both the geographic layout and ridership distribution.

\noindent The global and polygon-wise frameworks were compared by evaluating performance across all train–test pairs (Tables~\ref{tab:polygon_id_features_results} and~\ref{tab:global_id_features_results}). To assess whether the polygon-based approach outperforms the global approach under both feature regimes, we applied  complementary paired tests across 14 matched (month, week) sets, using the Wilcoxon signed-rank test. Results are reported for models trained with ID features (Table~\ref{tab:polygon_vs_global_mae_wilcoxon_compact}) and without ID features (Table~\ref{tab:polygon_vs_global_mae_wilcoxon_compact_nonid}). In both feature settings, LightGBM and CatBoost, and in most cases Random Forest, show statistically significant improvements for the polygon-wise models ($p<0.05$), whereas XGBoost and Linear Regression do not.

\noindent Although the absolute MAE differences are small (typically $\leq 0.05$), their consistency across evaluation periods indicates a modest but reliable advantage for polygon-based modeling, independent of whether ID features are included. To assess the magnitude of this difference, we applied Cliff’s Delta ($\delta$) effect size analysis. For the high-performing tree-based models, the effect size remains in the “small” to “negligible” range (e.g., $\delta = 0.194$ for LightGBM and $\delta = 0.133$ for CatBoost in the non-ID regime), confirming that while the polygon-wise framework provides a statistically significant edge, the per–test set improvement is subtle in magnitude.

\begin{figure*}[ht]
     \centering
     \includegraphics[width=1\textwidth]{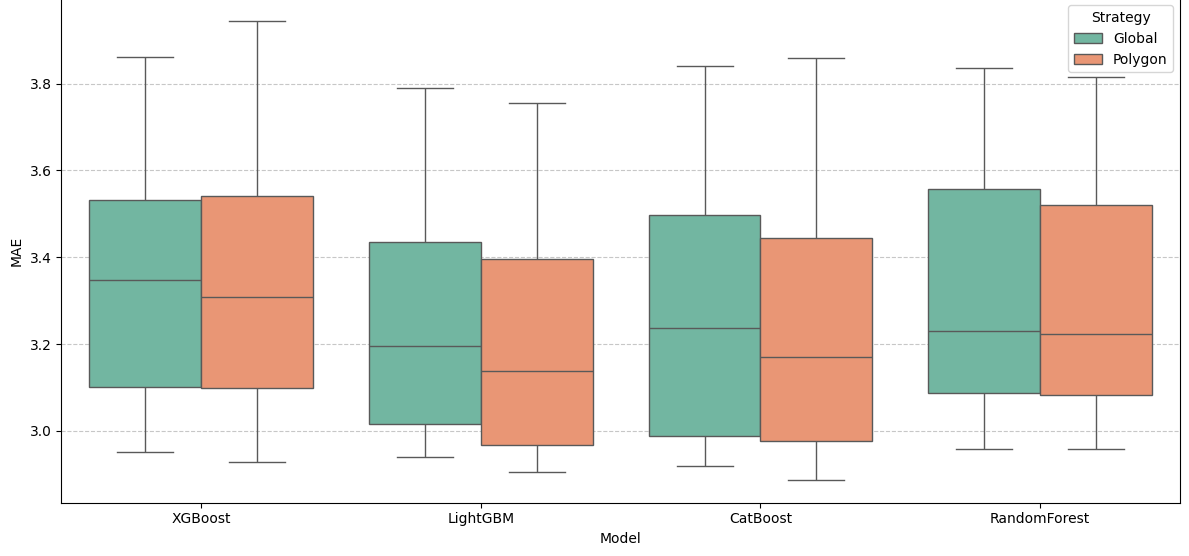}
     \caption{Distribution of mean absolute error (MAE) for four tree-based models (XGBoost, LightGBM, CatBoost, and Random Forest) under the global and polygon-based strategies, for all the test sets used in the experiment. The boxplots show that both approaches yield comparable error distributions, with the polygon-based models generally achieving slightly lower median MAE values.}
     \label{fig:MAE_box_plot_polygon_vs_global}
\end{figure*}

\noindent Including ID features did not yield systematic improvements (Figures~\ref{fig:MAE_box_plot_polygon_vs_global} and~\ref{fig:RMSE_global_polygon_id_no_id}); the error distributions with and without IDs are nearly indistinguishable, suggesting that generalizable features provide most of the predictive signal. Nevertheless, paired significance tests indicate that small but consistent differences do exist between the with-ID and without-ID settings for some models: LightGBM benefits modestly from IDs in both polygon and global configurations, and XGBoost shows a minor effect in the polygon setting, whereas CatBoost and Random Forest exhibit no reliable gains (Tables~\ref{tab:wilcoxon_polygon_id_vs_noid_compact} and~\ref{tab:wilcoxon_global_id_vs_noid_compact}). 

\noindent Among all evaluated models, LightGBM consistently achieved the lowest error rates on the test sets, with an average MAE of 3.25 across all 14 test sets (as shown in Tables~\ref{tab:polygon_id_features_results} and~\ref{tab:global_id_features_results}). It marginally outperformed the other tree-based models. Therefore, the polygon-wise LightGBM model, which also recorded the lowest average mean absolute error (MAE) across test sets, was selected for further analysis. Its predictive performance was evaluated in comparison with its global counterpart across temporal dimensions and ridership levels. The SHAP analysis of feature importance was conducted solely on the polygon-wise model.

\begin{figure}[ht]
     \centering
     \includegraphics[width=1\linewidth]{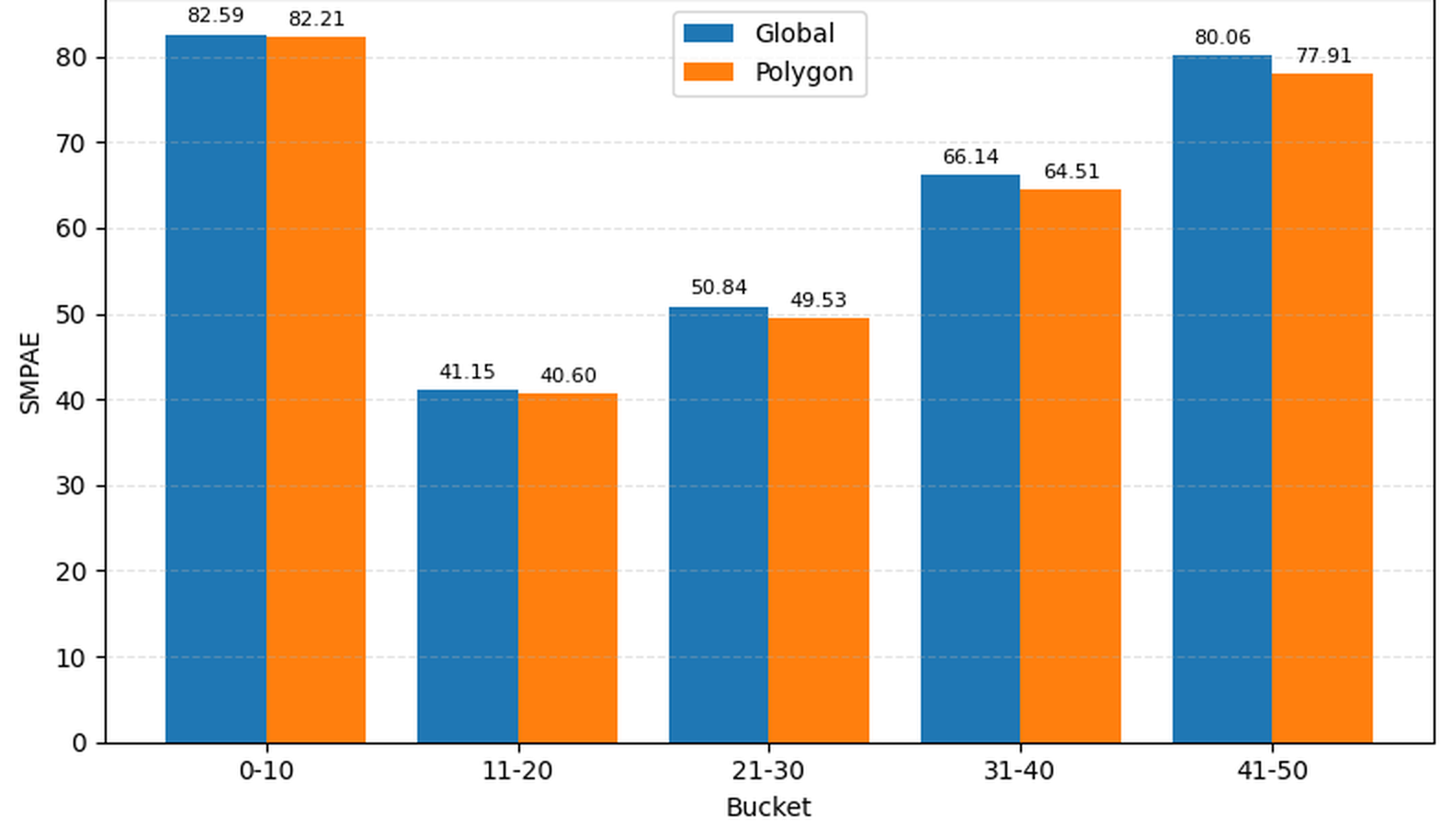}
     \caption{sMPAE of LightGBM across ridership buckets (0–10 to 41–50) over all experiments, comparing global and polygon strategies. Both approaches show similar performance, with the highest errors in the lowest ridership bucket and a monotone increase in error from 11–20 up to 41–50 ridership.}
     \label{fig:lightgbm_SMPAE_bucket}
\end{figure}

\noindent Evaluation by ridership level (Figure~\ref{fig:lightgbm_SMPAE_bucket}) indicates that percentage errors are highest in the lowest ridership bucket (0–10 passengers) and increase gradually with higher passenger volumes.  Hourly patterns (Figures~\ref{fig:lightgbm_RMSE_p_hourly} and \ref{fig:lightGBM_MAE_hourly}) show lower errors during most of the day, with higher variance at night and during peak hours.

\noindent Figure~\ref{fig:lightgmb_MAE_week} shows that extending the LightGBM training set to include the second-to-last week did not yield meaningful improvements in predictive accuracy; although the polygon-wise variant recorded a slightly lower median error in the final week, the difference was marginal. Figure~\ref{fig:lightgbm_month_sMPAE} displays the average sMPAE for both polygon-wise and global LightGBM models. Models performance remained consistent throughout the year, with differences typically within one percentage point. The highest error was observed in April at approximately 80\%, while the lowest occurred in May at around 64\%.

\begin{figure*}[ht]
  \centering
  \begin{subfigure}[t]{0.435\textwidth}
    \centering
    \includegraphics[width=\linewidth, height = 0.27\textheight]{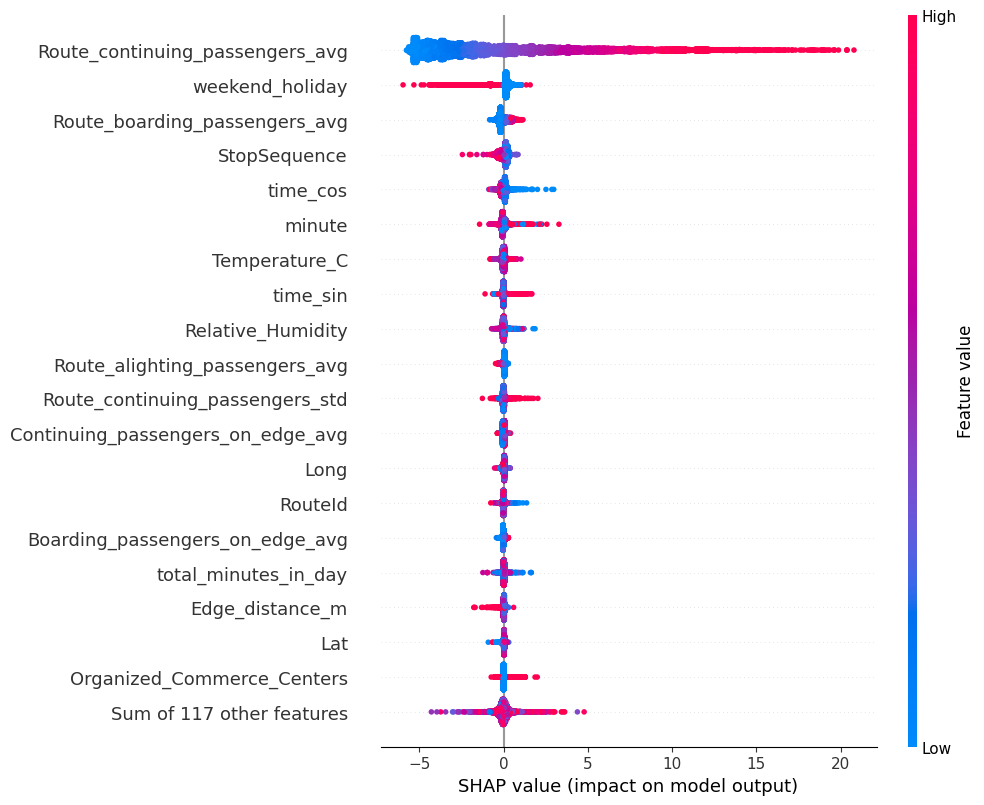}
    \caption{January with identifier features}
    \label{fig:shap_jan_id}
  \end{subfigure}
  \hfill
  \begin{subfigure}[t]{0.48\textwidth}
    \centering
    \includegraphics[width=\linewidth, height = 0.27\textheight]{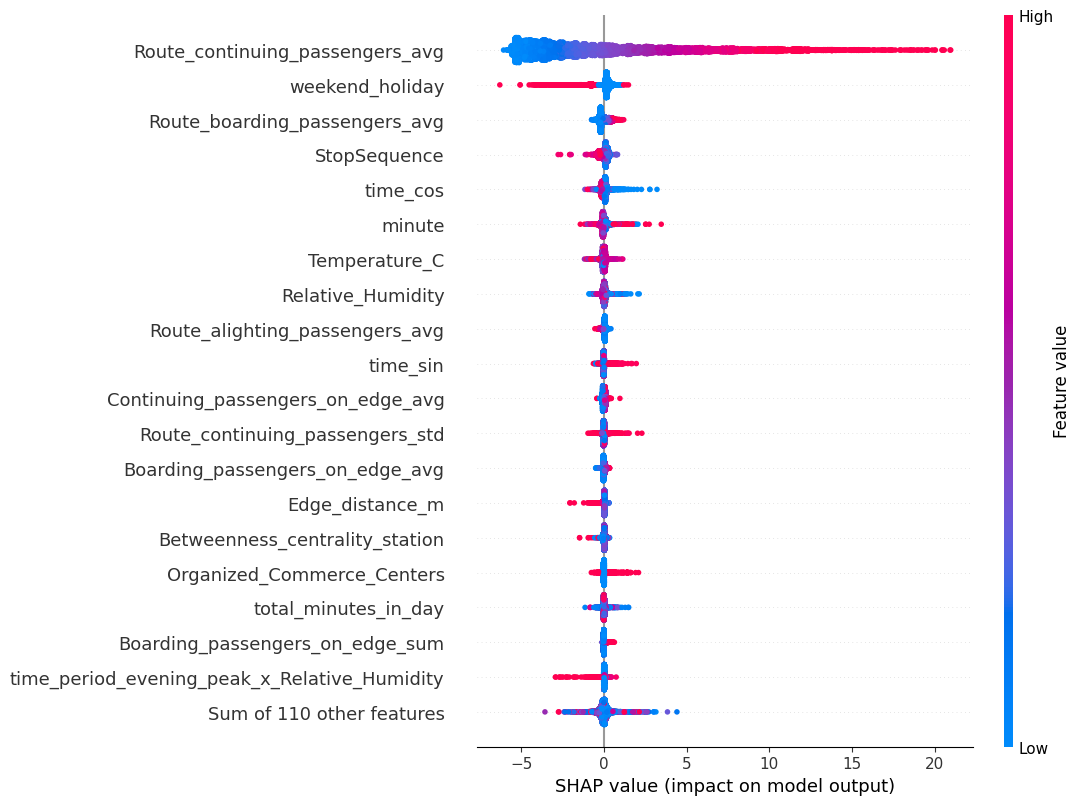}
    \caption{January without identifier features}
    \label{fig:shap_jan_noid}
  \end{subfigure}
  \caption{SHAP values for January (polygon-level LightGBM).}
  \label{fig:shap_january}
\end{figure*}

\noindent SHAP analysis of the polygon-wise LightGBM model (Figures~\ref{fig:shap_jan_id} and \ref{fig:shap_may_noid}) provides transparency into the model's decision-making process.\newline The results identify \texttt{Route\allowbreak\_continuing\allowbreak\_passengers\allowbreak\_avg} as the most influential feature by a significant margin. This suggests that the historical occupancy baseline of a specific bus line is the primary anchor for real-time predictions. Other key insights from the SHAP values include:
\begin{itemize}[leftmargin=*]

    \item \textbf{Historical and Spatial Baselines:} Beyond route-wide averages, the model relies on segment-specific data such as 
    \texttt{Continuing\_passengers\_}\allowbreak\texttt{on\_edge\_avg} and \texttt{Boarding\_passengers\_}\allowbreak\texttt{on\_edge\_avg}. These features allow the model to adjust predictions based on localized demand hot spots that may be shared across different routes.
    
    \item \textbf{Temporal Dynamics:} The high ranking of \newline \texttt{weekend\_holiday}, \texttt{time\_cos}, and \newline \texttt{total\_minutes\_}\allowbreak\texttt{in\_day} underscores the strong cyclical nature of bus ridership. The model effectively captures the variance between peak-hour surges and reduced holiday demand, with \texttt{time\_cos} and \texttt{time\_sin} providing a continuous representation of 24-hour patterns.
    
    \item \textbf{Environmental and Contextual Factors:}
    \newline\texttt{Temperature\_C} and \texttt{Relative\_Humidity} show a measurable impact, indicating that weather conditions influence passenger choice. Furthermore, the inclusion of \texttt{Organized\_}\allowbreak\texttt{Commerce\_Centers} as a spatial feature highlights how proximity to specific land-use types serves as a proxy for trip generation.
    
    \item \textbf{Network Topology:} Graph-based indicators, including \texttt{StopSequence} and geographical coordinates (\texttt{Lat}, \texttt{Long}), contribute to the model's spatial awareness, allowing it to account for the distance-based decay of ridership and the stop's position within the route hierarchy.
    
\end{itemize}

\noindent Collectively, these SHAP values demonstrate that while historical ridership is the strongest predictor, the model achieves high accuracy by layering temporal cycles and localized urban context on top of these baselines.

\begin{table*}[t]
    \centering
    \caption{Average Performance of Polygon Models with ID Features (Based on 14 Experiments)}
    \textbf{Note}: Best values for each metric are bolded.
    \label{tab:polygon_id_features_results}

    \begingroup
    \setlength{\tabcolsep}{8pt} 
    \renewcommand{\arraystretch}{1.2}

    \begin{tabular}{l ccccc}
        \toprule
        \textbf{Model} & \textbf{RMSE} & \textbf{MAE} & \textbf{MPAE} & \textbf{\%RMSE} & \textbf{SMPAE} \\
        \midrule
        \textbf{CatBoost}         & 4.87 & 3.25 & 2.75e+09 & 75.46 & 75.60 \\
        \textbf{LightGBM}         & \textbf{4.81} & \textbf{3.21} & 2.79e+09 & \textbf{74.48} & \textbf{74.21} \\
        \textbf{LinearRegression} & 2002.89 & 50.37 & 1.12e+11 & 3.18e+04 & 177.34 \\
        \textbf{RandomForest}      & 4.96 & 3.30 & \textbf{2.74e+09} & 76.85 & 75.80 \\
        \textbf{XGBoost}          & 5.02 & 3.36 & 2.82e+09 & 77.90 & 77.38 \\
        \bottomrule
    \end{tabular}
    \endgroup
    \vspace{0.35em}
\end{table*}

\begin{table*}[t]
    \centering
    \caption{Average Performance of Global Models with ID Features (Based on 14 Experiments)}
    \textbf{Note}: Best values for each metric are bolded.
    \label{tab:global_id_features_results}

    \begingroup
    \setlength{\tabcolsep}{8pt}
    \renewcommand{\arraystretch}{1.2}

    \begin{tabular}{l ccccc}
        \toprule
        \textbf{Model} & \textbf{RMSE} & \textbf{MAE} & \textbf{MPAE} & \textbf{\%RMSE} & \textbf{SMPAE} \\
        \midrule
        \textbf{CatBoost}         & 4.90 & 3.28 & 2.85e+09 & 75.89 & 76.36 \\
        \textbf{LightGBM}         & \textbf{4.86} & \textbf{3.25} & 2.91e+09 & \textbf{75.29} & \textbf{74.67} \\
        \textbf{LinearRegression} & 525.45 & 26.27 & 4.41e+10 & 8.32e+03 & 78.00 \\
        \textbf{RandomForest}      & 4.98 & 3.32 & \textbf{2.76e+09} & 77.14 & 75.93 \\
        \textbf{XGBoost}          & 5.02 & 3.38 & 2.96e+09 & 77.85 & 77.67 \\
        \bottomrule
    \end{tabular}
    \endgroup
    \vspace{0.15em}
\end{table*}

\section{Discussion}
\label{sec:discussion}
Building on the results presented in Section~5, several key observations can be drawn:

\noindent First, Tables~\ref{tab:polygon_id_features_results} and \ref{tab:global_id_features_results} demonstrate that tree-based models significantly outperformed the linear regression models. The global linear regression model performed similarly to the tree-based models on the training set, achieving an MAE of 3.01, while the tree-based models (XGBoost, CatBoost, and LightGBM) 
ranged between 2.0 and 3.0 MAE. However, the polygon-wise LightGBM achieved the best overall performance across all test sets, with an average MAE of 3.21, compared to the simple global linear regression model, which recorded an MAE of 26.27 and thus failed to capture the underlying 
ridership patterns. The polygon-wise linear regression model showed a similar pattern, with a training MAE of 2.96 but a test MAE of 50.1, indicating extremely poor generalization ability and highlighting the highly non-linear nature of the underlying data.

\noindent Second, although the numerical differences in MAE between the polygon-based and global models are small (0.01–0.05) as seen in Figure~\ref{fig:MAE_box_plot_polygon_vs_global}, the paired Wilcoxon signed-rank tests (Tables~\ref{tab:polygon_vs_global_mae_wilcoxon_compact} and~\ref{tab:polygon_vs_global_mae_wilcoxon_compact_nonid}) consistently flag these gaps as statistically significant for the tree-based models. Because Wilcoxon is rank-based and robust to mild deviations from normality, the significance chiefly reflects the consistency of the polygon model’s advantage across the 14 matched (month, week) pairs rather than large effect sizes in any single period. In practice, this points to a stable, systematic, albeit modest, benefit of localized modeling that persists across evaluation windows. This modest scale is further confirmed by the Cliff's Delta ($\delta$) results, which categorize the effect sizes for the tree-based models as small to negligible (ranging from 0.02 to 0.19), reinforcing that the framework provides a reliable but subtle improvement in predictive performance.

\noindent Turning to the role of ID features, the paired Wilcoxon tests comparing with-ID vs. {without-ID settings show only limited, model-specific gains (Figure~\ref{fig:RMSE_global_polygon_id_no_id}; Tables~\ref{tab:wilcoxon_polygon_id_vs_noid_compact} and~\ref{tab:wilcoxon_global_id_vs_noid_compact}) show that that are statistical diffrences between the settings. In the polygon setting, LightGBM exhibits a small but significant improvement with IDs (mean $\Delta$MAE $\approx -0.02$), and XGBoost shows a similar pattern. CatBoost and Random Forest show no reliable change. In the global setting, LightGBM again benefits modestly (mean $\Delta$MAE $\approx -0.04$), whereas CatBoost, Random Forest, and XGBoost remain statistically unchanged. Linear Regression displays a large, significant gap in the global setting, indicating strong reliance on identifiers and instability rather than a generalizable benefit of IDs. 

\noindent Since ID features are strong predictors, some degradation when they are removed is expected. The fact that performance remains very close across the two feature regimes is therefore encouraging: it suggests that most predictive power comes from generalizable, system-agnostic features, and it supports optimism about the models’ transferability to settings where identifiers differ or evolve.

\noindent Third, the results in Tables~\ref{tab:polygon_id_features_results} and \ref{tab:global_id_features_results} demonstrated that polygon-wise models achieved performance levels comparable to, and in some cases slightly exceeding, those of a single global model. This outcome indicates that spatial partitioning provided only a modest advantage. However, the utility of the polygon-wise approach may be significantly more pronounced in larger metropolitan areas, such as the Tel Aviv metropolitan region. In such expansive systems, the high degree of spatial non-stationarity and the existence of multiple, distinct functional centers would likely benefit more from the localized learning offered by the max-p regionalization framework.

\noindent Fourth, tree-based models produced highly similar results (Tables~\ref{tab:polygon_id_features_results} and \ref{tab:global_id_features_results}), with only minor differences observed across experiments. Random Forest, despite strong performance on training data, exhibited clear overfitting on test sets, which highlighted its limited generalization capability. Overall, tree-based methods generalized more effectively than linear models. Among these, LightGBM achieved the most consistent and highest performance in terms of mean absolute error (MAE) and was therefore selected, in both polygon-wise and global variants, for further evaluation under varying conditions.

\noindent Fifth, evaluation of model performance under temporal conditions revealed two key patterns. At the seasonal scale, monthly results indicated that polygon-wise and global LightGBM models performed nearly identically throughout the year, with differences typically within one percentage point in percentage-based metrics. To complement this, Figure~\ref{fig:lightGBM_MAE_month} presents the Mean Absolute Error (MAE) across months, providing an absolute-scale perspective on prediction accuracy. While percentage-based metrics such as sMAPE and \%RMSE capture relative error magnitudes, MAE offers an interpretable measure of absolute deviation that is unaffected by small denominators or near-zero values. The results confirm that both models exhibit remarkably stable absolute errors, ranging from approximately 3.0 passengers in winter months to about 3.6 passengers in April. The lowest MAE values were recorded in May (Figure~\ref{fig:lightGBM_MAE_month}), while April showed the highest, consistent with the percentage-based findings. This temporary increase is likely attributable to the disruptive effect of the Pesach holiday, during which ridership deviated sharply from regular patterns. Overall, the MAE analysis reinforces that seasonal fluctuations in predictive accuracy are minimal and that both modeling strategies maintain consistent performance throughout the year, demonstrating their robustness under typical operating conditions.

\noindent At the daily scale,  Figure~\ref{fig:lightgbm_RMSE_p_hourly} illustrates that the highest percentage based errors occurred during the night and early morning hours. These periods correspond to times of low and highly inconsistent ridership. The results indicate that small absolute deviations during low-demand periods resulted in disproportionately large percentage errors. This phenomenon complicates accurate forecasting. To complement these findings, Figure~\ref{fig:lightGBM_MAE_hourly} presents the Mean Absolute Error (MAE) by hour, offering an absolute-scale perspective that is unaffected by small denominators. The MAE values closely follow the diurnal ridership cycle and increase sharply during the morning peak.
The morning peak occurs around 07:00 to 09:00, with maximum deviations near midday, approximately 13:00 to 15:00, before gradually declining in the evening. Absolute errors remain low during the night and early morning, indicating that large percentage errors in those periods are primarily due to very small denominators rather than poor predictive accuracy. Polygon-wise and global LightGBM models exhibited nearly identical MAE patterns across all hours. Differences were typically below 0.2 passengers, confirming that both approaches provide comparable temporal stability.

\noindent Sixth, after evaluating the influence of temporal and seasonal factors, predictive accuracy was 
further analyzed across varying ridership levels using LightGBM models. As figure ~\ref{fig:lightGBM_RMSE_buckets} illustrates, the largest errors occurred in the lowest ridership bucket (0–10 passengers), where both polygon-wise and global variants yielded Percent RMSE values above 96\%. This outcome demonstrates the instability of percentage-based 
metrics under very low demand, as minor absolute deviations can result in disproportionately large percentage errors. To complement this, Figure~\ref{fig:lightGBM_MAE_buckets} presents the MAE across the same ridership buckets. In contrast to the percentage-based results, MAE values increased almost linearly with ridership volume, ranging from approximately 2.4 passengers for the lowest bucket to around 24 passengers for the highest (41–50) bucket. The absolute deviations therefore scale naturally with passenger volume, providing a more interpretable sense of magnitude than percentage errors. The differences between polygon-wise and global strategies were minimal, 
typically below 0.3 passengers across all buckets, confirming that both modeling strategies deliver 
comparable predictive accuracy across ridership conditions. Combined with the temporal analysis, these results indicate that the most significant challenges for accurate prediction occur under sparse or irregular demand conditions, such as late at night, during holiday periods, or within the lowest ridership buckets. 

\noindent Seventh, to assess whether expanding the training set with additional data improves predictive 
accuracy, Figure~\ref{fig:lightgmb_MAE_week} illustrates the MAE values obtained under the iterative evaluation framework, where predictions for the final test week of each month were generated using training data that included observations from the preceding (second to last) week. The results indicate that including the additional week in training did not lead to a measurable improvement in predictive performance. Both the polygon-wise and global LightGBM models exhibited similar MAE distributions across the two test weeks, with median errors ranging between approximately 3.0 and 3.4 passengers. Although the polygon-wise model achieved slightly lower median errors in both weeks, the differences were minor. The stability of MAE across weeks suggests that the temporal variability of ridership patterns limits the benefit of simple training set extensions, highlighting the need for more adaptive temporal features or dynamic learning mechanisms to improve generalization.

\noindent Eight, the relationship between model accuracy and transit volume, as illustrated in the Error vs. Demand Envelope (Figure~\ref{fig:station_mae_lineraity}), reveals a highly linear progression ($R^{2} = 0.904$). This suggests that the Mean Absolute Error (MAE) is primarily a function of the scale of ridership; as the number of passengers increases, the absolute variance and subsequent prediction difficulty grow proportionally. However, the correlation analysis of absolute error against land-use types provides a deeper layer of explanation. Table~\ref{tab:poi_error_corr} highlights that the highest error correlations are associated with the Old City (0.105) and Civic Center (0.097) zones. Unlike residential or commercial areas, these hubs often host non-routine events, tourism, and irregular administrative activities that introduce stochastic noise into the demand profile. This explains why, despite the overall linear trend, certain stations remain more challenging to predict; their difficulty is rooted in the complex, non-periodic human behavior associated with specialized urban functions.

\begin{table*}[ht]
\centering
\caption{Correlation between selected land-use features and absolute prediction error}
\label{tab:poi_error_corr}
\begin{tabular}{l c}
\hline
\textbf{Feature} & \textbf{Correlation with $|$Error$|$} \\
\hline
Old City & 0.105 \\
Civic Center & 0.097 \\
Hospital & 0.083 \\
Neighborhood Shopping Centers & 0.073 \\
Education & 0.067 \\
Health & 0.066 \\
Daycares & 0.064 \\
University & 0.062 \\
Street-Accompanied Commerce & 0.061 \\
Organized Commerce Centers & 0.060 \\
\hline
\end{tabular}
\end{table*}

\noindent Last point regarding the results of the experiments is, that after evaluating performance under varying temporal and ridership conditions, the drivers of model predictions were examined using SHAP (Figures~\ref{fig:shap_jan_id} and \ref{fig:shap_may}). The analysis identified a clear hierarchy among predictors. The most influential feature was the average ridership carried forward from previous stops along the route, which strongly shaped model outputs and was consistent with expectations. Additional ridership dynamics, such as the average number of boardings per hour on a route segment, also ranked highly. Temporal variables, including sine and cosine time transformations, weekday versus weekend indicators, and stop sequence within the trip, contributed substantially to predictive accuracy.

\noindent Weather effects were evident, with temperature and relative humidity consistently ranking high, while rainfall had little influence. This likely reflects Be’er Sheva’s low annual precipitation. In contrast, graph-based and land-use features played a more limited role. Graph measures such as station-level betweenness centrality, edge-level passenger flow statistics, and inter-stop distances appeared among the top 20 predictors in January but contributed less overall. Similarly, land-use variables, particularly the number of organized commercial centers near a stop, were occasionally influential but had a smaller impact compared to ridership and temporal drivers.

\noindent A primary concern during the preprocessing stage was the quality and reliability of the ridership data, which were occasionally compromised by inaccuracies in the APC system’s laser-based measurements of boardings and alightings. These sensor-based errors introduced inconsistencies in the dataset, such as implausible negative or abnormally large ridership values, necessitating additional validation and filtering procedures before model training.

\noindent To further assess data consistency, APC-derived ridership values were cross-compared with corresponding records from the AV system, yielding a Pearson correlation coefficient of approximately 0.6. This indicates a moderate level of agreement between the two data sources. In addition, supplementary analyses were conducted using the AVL dataset, examining correlations across different ridership levels to verify the robustness and stability of these patterns. Additionally, the coefficient of variation (CV) of ridership across routes was computed separately for the training and testing sets within hourly time intervals throughout the day (Figure~\ref{fig:CV_heatmap}). The CV values ranged from approximately 40\% to 160\%, with warmer colors in the heatmap representing greater variability. Further details and extended diagnostic results are provided in Appendix ~\ref{app:APC limitation}.

\noindent This research makes two primary contributions to the field of PT demand forecasting. First, it introduces a \textbf{novel methodological framework} by integrating \textit{max-p regions} with tree-based machine learning ensembles. To the best of the our knowledge, this specific data-driven partitioning framework has not been previously implemented to address spatial non-stationarity in transit ridership.

\noindent Second, this work provides a \textbf{theoretical validation of Tobler’s First Law of Geography} within an operational transit context. By demonstrating that localized polygon-wise models can achieve high predictive accuracy (LightGBM $\delta = 0.184$, CatBoost $\delta = 0.133$ in non-ID regimes), the study confirms that transit demand drivers are not spatially uniform. This suggests that "near things" within a specific urban cluster share more common predictive characteristics than the network as a whole. Consequently, this study challenges the reliance on global "black-box" averages, providing a geographically-sensitive alternative that respects the inherent heterogeneity of the urban environment.

\section{Future Work}
\label{sec:Future Work}
A comprehensive framework was developed and evaluated for predicting bus ridership at the stop level in Be'er Sheva, integrating machine learning methods with graph-based spatial representations. Systematic comparison of global and polygon-wise modeling strategies demonstrated that geographically localized models can achieve accuracy similar to global models, while substantially outperforming simpler methods such as linear regression. The results also highlighted the robustness of tree-based algorithms, particularly LightGBM, in capturing complex nonlinear ridership dynamics. Based on these findings, several promising directions for future research are identified.

\noindent One potential area involves implementing advanced geographically aware models. Techniques such as Geographically Weighted Regression (GWR), Geographically Weighted Random Forests, and spatially constrained gradient boosting may more effectively capture localized variations in ridership patterns and address spatial heterogeneity compared to global models.

\noindent A further research direction is the evaluation of alternative clustering methods for partitioning stations beyond the Max-p algorithm. Methods including SKATER, AZP, and community detection could yield spatial aggregations that more accurately reflect functional travel regions within the city. A systematic comparison of these clustering approaches would identify which partitioning strategies most effectively support localized prediction.

\noindent While this work focuses on bus ridership forecasting,
previous studies have explored various methods to improve scheduling:
\begin{itemize}
    \item \textbf{Minimizing costs and delays:} Optimization models and genetic algorithms have been used to reduce operational costs, passenger waiting times, and transfer times by incorporating factors such as stochastic delays and dynamic passenger behavior \cite{wu2015stochastic, wu2019stochastic, fonseca2018matheuristic, ma2020single, wang2024research}.
    \item \textbf{Enhancing robustness:} Researchers have developed models that account for travel-time uncertainties to create more robust and reliable timetables, mitigating the impact of disruptions \cite{yan2012robust, muller2022estimating}.
\end{itemize}

\noindent These areas represent promising avenues for future work, building on the foundation of accurate passenger flow prediction to create more efficient and resilient PT systems, especially by optimizing bus frequencies to match passenger demand.

\section{Conclusions}
\label{sec: Conclusions}
This study presents a machine-learning-based methodology to forecast PT ridership. Our framework follows a four-step process: data collection (see Section~\ref{subsec:Data Collection and Preprocessing}), feature extraction, graph analysis and polygon creation (see Section~\ref{subsec:Data Analysis}) and model evaluation and analysis (see Section ~\ref{subsec:ModelEvaluation}). The analysis compared linear and tree-based regression models under both global and polygon-wise configurations, with additional experiments examining the role of identifying (ID) features and the contribution of different feature groups. The tree-wise models consistently outperformed linear regression, demonstrating their superior ability to capture complex nonlinear relationships in the data. Although the polygon-wise models exhibited metric value ranges comparable to those of the global models, such as MAE, statistical testing across multiple test sets revealed significant differences, as polygon-wise models consistently achieved slightly better results. This indicates that, despite similar absolute performance levels, spatial partitioning provides measurable benefits by capturing localized behavioral patterns.

\noindent When comparing models trained with and without ID features, the results showed statistically significant differences, yet the overall performance metrics were nearly identical. Given that identifiers are expected to exert a strong influence, this similarity underscores that the feature engineering and data analysis process effectively captured the underlying patterns and trends in passenger behavior without relying on explicit system identifiers. The SHAP analysis revealed that the most influential characteristic by far was the average ridership on the route at each stop, followed by temporal indicators (such as hours and weekends / weekdays) and weather-related conditions. Importantly, meaningful predictors emerged from all feature domains, urban facilities, weather, and network structure, highlighting the value of integrating diverse data layers in the modeling pipeline.

\noindent Nevertheless, the models exhibited limitations in capturing ridership dynamics during non-rush hours, weekends, and holidays, reflecting the inherently irregular and less predictable nature of passenger flow in those periods. Furthermore, the inclusion of additional additive datasets did not lead to measurable performance gains, suggesting diminishing returns beyond the current integrated feature framework. Overall, the findings confirm the robustness of the proposed modeling approach, the relevance of spatial partitioning, and the importance of combining multiple contextual data sources for accurate and interpretable ridership prediction.

\phantomsection
\section*{Acknowledgments} 

\addcontentsline{toc}{section}{Acknowledgments} 
We thank the Israel Innovation Authority  for providing
the data for this study. In addition, while drafting this article, we used ChatGPT, Gemini and Grammarly for slight editing according to necessity.

\phantomsection
\bibliography{sample}

\section*{Appendices}
\addcontentsline{toc}{section}{Appendices}

\appendix
\renewcommand{\thesection}{\Alph{section}}

\renewcommand{\thefigure}{S\arabic{figure}}
\renewcommand{\thetable}{S\arabic{table}}
\setcounter{figure}{0}
\setcounter{table}{0}

\section{APC data limitation}\label{app:APC limitation}
Although cleaning and filtering procedures were implemented to mitigate these issues, the persistence of such errors reduces the reliability of the dataset and could compromise the accuracy of the results. This underscores the need for improved data collection and validation methods in subsequent research.

\noindent Figure~\ref{fig:hist_plot_bus_ridership} shows the distribution of bus ridership at stops across all observations. While most values are non-negative (blue), a substantial portion of negative values (red) is also present, which are not plausible in reality. Specifically, 103,651 out of 220,823 trips ($\sim$47\%) included at least one negative ridership measurement, highlighting the severity of this data quality issue and the need for rigorous data cleaning.

\noindent Beyond the presence of negative values, statistical methods were applied to evaluate the overall noise level in the ridership data. Figure~\ref{fig:CV_heatmap} illustrates that coefficients of variation (CV), defined as the ratio of the standard deviation to the mean,
\[
\text{CV} = \frac{\sigma}{\mu} \times 100,
\]  
remained elevated across routes, even after the data were divided into training and testing sets by time bucket for the last week of January. A high CV indicates that fluctuations in ridership counts are substantial relative to the mean, reflecting instability in the underlying measurements. Although the exclusion of weekends and holidays reduced calendar-driven variability, significant inconsistencies persisted, confirming that the ridership data are both noisy and unstable. The evening and late-night periods from 18:00 to 23:00 exhibited the highest variability, with CV values approaching 100 in both sets, which underscores particularly inconsistent demand during these hours.

\noindent Robustness statistics further corroborate this finding. Table~\ref{tab:train_test_january} shows that the ratios of median absolute deviation (MAD) to median (0.75) and interquartile range (IQR) to median (2.0) were consistent across both datasets. These results confirm the presence of numerous outliers and substantial deviations from the median, underscoring the high variability and inconsistency in the ridership data. This variability also appeared as systematic differences between the training and testing datasets. Significant discrepancies persisted even after controlling for route, direction, departure time, weekday or weekend status, and stop. For the last week of January, the absolute relative difference in average ridership between the two sets was approximately 44\%, indicating that the training data does not reliably represent the testing conditions and that the two datasets reflect distributions with distinct underlying characteristics.

\begin{table}[t]
\centering
\caption{Robustness statistics (MAD/Median and IQR/Median) for the train–test split of the last week of January. Lower values indicate greater robustness.}
\label{tab:train_test_january}
\begin{tabularx}{\linewidth}{l>{\centering\arraybackslash}X>{\centering\arraybackslash}X}
\toprule
\textbf{Dataset} & \textbf{MAD/Median} & \textbf{IQR/Median} \\
\midrule
Train & 0.75 & 2.0 \\
Test  & 0.75 & 2.0 \\
\bottomrule
\end{tabularx}
\end{table}

\begin{table}[!ht] 
  \centering 
  \small 
  
  \caption{Model Performance: Error Metrics (May)}
  \label{tab:performance_errors}
  \begin{tabular*}{\columnwidth}{@{\extracolsep{\fill}} c r r r r @{}}
    \toprule
    \textbf{Bucket} & \textbf{n} & \textbf{MAE} & \textbf{RMSE} & \textbf{MAPE} \\
    \midrule
    0--10  & 314,751 & 0.81 & 1.52 & 0.45 \\
    11--20 &   2,498 & 5.12 & 7.24 & 0.37 \\
    21--30 &     242 & 17.66 & 19.88 & 0.72 \\
    31--40 &      81 & 28.78 & 30.71 & 0.83 \\
    41--50 &      33 & \textbf{40.94} & \textbf{41.55} & \textbf{0.93} \\
    \bottomrule
  \end{tabular*}

  \vspace{0.5cm} 

  \caption{Model Performance: Volume \& Mean (May)}
  \label{tab:performance_means}
  \begin{tabular*}{\columnwidth}{@{\extracolsep{\fill}} c r r r @{}}
    \toprule
    \textbf{Bucket} & \textbf{WAPE} & \textbf{AVL Mean} & \textbf{APC Mean} \\
    \midrule
    0--10  & 0.52 & 0.39 & 2.07 \\
    11--20 & 0.53 & 13.33 & 11.11 \\
    21--30 & 1.30 & 24.21 & 9.24 \\
    31--40 & 1.57 & 34.40 & 6.75 \\
    41--50 & \textbf{1.78} & 44.15 & 3.64 \\
    \bottomrule
  \end{tabular*}
\end{table}

\noindent To further assess data reliability, we compared APC onboarding counts with the AVL (Rav-Kav) dataset for May, which contains validated boardings. After merging at the trip level, 317,605 trips were analyzed using correlation and error metrics, along with bucket-based evaluations consistent with the main procedure. The correlations (Pearson’s $r$ = 0.661, Spearman’s $\rho$ = 0.634, Kendall’s $\tau$ = 0.567, CCC = 0.656; all $p < 0.001$) indicate a moderate relationship between the two sources, confirming that APC counts are not random but still subject to substantial discrepancies.

\noindent To assess the extent of these discrepancies, we evaluated the precision of the APC in the passenger groups (Table~\ref{tab:performance_errors}). For low-ridership trips (0–10 boardings, $n=314{,}751$), which comprised the majority of samples, the APC system aligned reasonably well with Rav-Kav (MAE = 0.81, RMSE = 1.52). Accuracy declined sharply at higher ridership levels: in the 11–20 range, MAE rose to 5.1, and by 21–30 it reached 17.7. The largest errors occurred in the 41–50 range ($n=33$), where Rav-Kav reported an average of 44 boardings but the APC system recorded only 3.6 (MAE = 40.9, RMSE = 41.6). These results indicate that while APC counts are reliable under low-demand conditions, they systematically undercount passengers on crowded trips, as reflected in very high relative error metrics (MAPE, sMAPE, and SMAPE > 90\% in the highest ridership categories).

\noindent Prior research has shown that real-world APC deployments often encounter situations that are difficult to classify accurately, resulting in measurement imprecision that must be statistically validated against acceptable bounds ~\citet{siebert2020validation}. This implies that systematic errors in passenger counts are expected, and careful validation procedures are essential when using APC data for research or operational decision-making. 

\noindent Similarly, ~\citet{berrebi2022cross} highlight that data quality issues consistently affect APC-based ridership analysis. Their research recommends cross-verifying APC counts with independent data sources to ensure consistency and completeness. Consistent with these findings, the inconsistencies identified in the Be’er Sheva APC dataset, including implausible negative values and discrepancies with AVL records, underscore the necessity for careful interpretation of model outcomes.

\begin{figure}[ht]
     \centering
     \includegraphics[width=1\linewidth, height = 0.3\textheight]{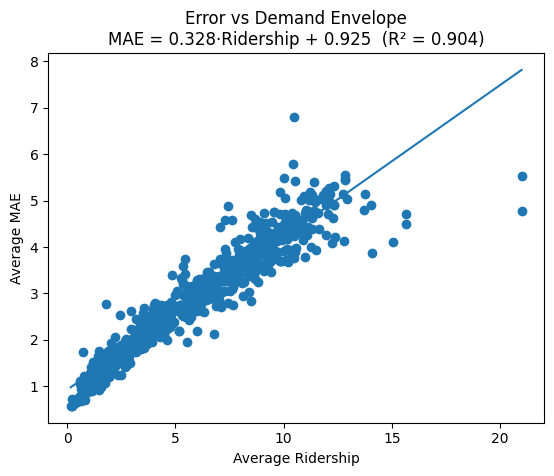}
     \caption{A scatter plot illustrating the strong linear correlation between average stop-level ridership and Mean Absolute Error (MAE). The regression model ($MAE = 0.328 \cdot Ridership + 0.925$) with a high $R^2 = 0.904$ indicates that the scale of demand is the primary driver of absolute prediction error.}
    \label{fig:station_mae_lineraity}
\end{figure}

\begin{figure*}[ht]
    \centering
    \includegraphics[width=1\linewidth, height = 0.3\textheight]{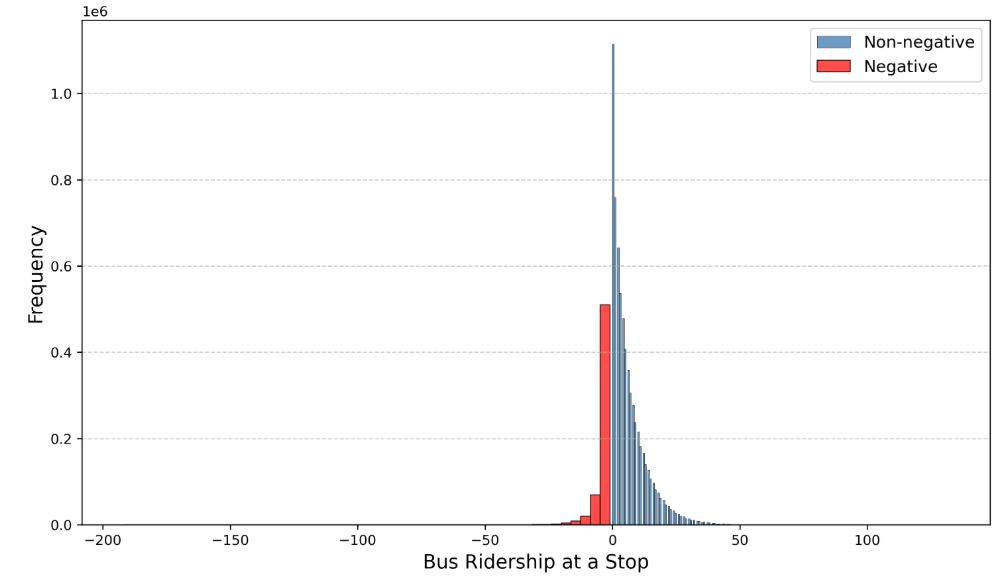}
    \caption{Histogram of bus ridership per stop, showing the distribution of recorded passenger counts across all observations. Bars in red indicate negative values, while bars in blue represent positive ridership after departing from a stop}
    \label{fig:hist_plot_bus_ridership}
\end{figure*}

\begin{figure*}[ht]
     \centering
     \includegraphics[width=1\linewidth, height = 0.4\textheight]{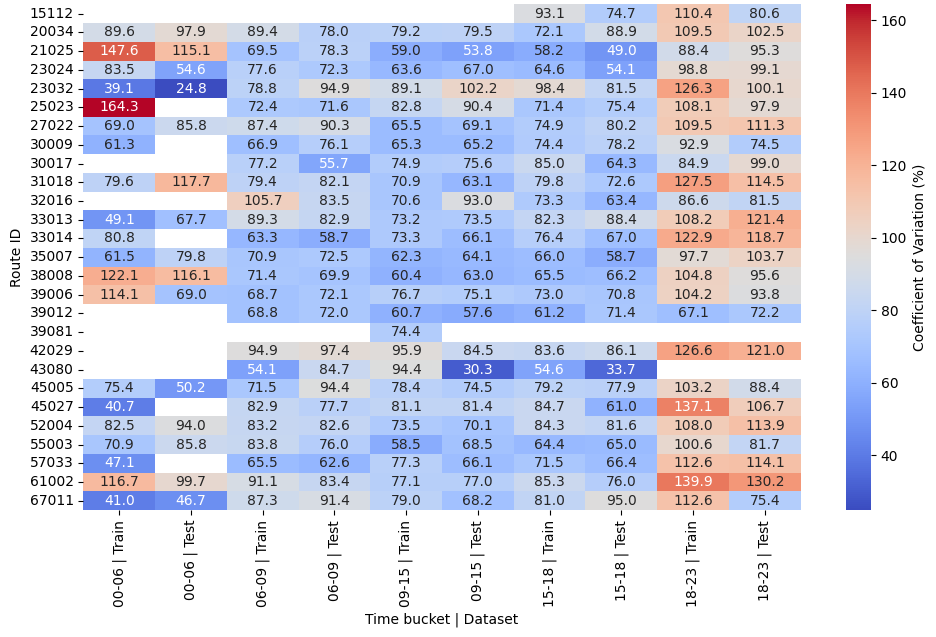}
     \caption{Coefficient of variation (CV) of ridership across routes, shown separately for training and testing datasets within each time bucket during the last week of January (excluding weekends and holidays). Warmer colors indicate higher variability.}
     \label{fig:CV_heatmap}
\end{figure*}

\clearpage

\onecolumn
\section{Supplementary Figures and Tables}\label{app:suppfigs}

\begin{table*}[ht]
\centering
\caption{Wilcoxon Signed-Rank Test Results: Polygon vs. Global MAE Differences Across 14 Matched Test Sets}
\textbf{Note}: Negative differences indicate lower polygon MAE. Statistically significant $p$-values ($<0.05$) are bolded. Cliff's Delta ($\delta$) indicates the effect size.
\label{tab:polygon_vs_global_mae_wilcoxon_compact}
\renewcommand{\arraystretch}{1.2}
\begin{tabularx}{\textwidth}{l *{5}{>{\centering\arraybackslash}X}}
    \toprule
    \textbf{Model} & 
    \shortstack{Mean \\ Diff \\ (MAE)} & 
    \shortstack{Median \\ Diff \\ (MAE)} & 
    \shortstack{Wilcoxon \\ Stat} & 
    \shortstack{$p$-\\Value} &
    \shortstack{Cliff's \\ Delta ($\delta$)} \\
    \midrule
    \textbf{LightGBM}          & -0.047 & -0.044 & 0.00  & \textbf{0.00} & 0.18  \\
    \textbf{CatBoost}          & -0.035 & -0.037 & 11.00 & \textbf{0.00} & 0.10  \\
    \textbf{RandomForest}      & -0.013 & -0.006 & 21.00 & \textbf{0.02} & 0.04  \\
    \textbf{XGBoost}           & -0.023 &  0.001 & 45.00 & 0.33          & 0.03 \\
    \textbf{LinearRegression}  &  24.11 &  19.08 & 89.00 & 0.99          & -0.45 \\
    \bottomrule
\end{tabularx}
\vspace{-25pt}
\begin{flushleft}
\end{flushleft}
\end{table*}

\begin{table*}[ht]
    \centering
    \caption{Wilcoxon Signed-Rank Test (Non-ID Features): Polygon vs. Global MAE Differences Across 14 Matched Test Sets}
    \textbf{Note}: Negative differences indicate lower polygon MAE. Statistically significant $p$-values ($<0.05$) are bolded. Cliff's Delta ($\delta$) indicates the effect size.
    \label{tab:polygon_vs_global_mae_wilcoxon_compact_nonid}
    \renewcommand{\arraystretch}{1.2}
    \begin{tabularx}{\textwidth}{l *{5}{>{\centering\arraybackslash}X}}
        \toprule
        \textbf{Model} & 
        \shortstack{Mean \\ Diff \\ (MAE)} & 
        \shortstack{Median \\ Diff  \\ (MAE)} & 
        \shortstack{Wilcoxon \\ Stat} & 
        \shortstack{$p$-\\Value} &
        \shortstack{Cliff's \\ Delta ($\delta$)} \\
        \midrule
        \textbf{LightGBM}          & -0.055 & -0.050 & 0.00  & \textbf{0.00} & 0.194  \\
        \textbf{CatBoost}          & -0.048 & -0.047 & 0.00  & \textbf{0.00} & 0.133  \\
        \textbf{RandomForest}      & -0.013 & -0.006 & 31.00 & \textbf{0.04} & 0.020  \\
        \textbf{XGBoost}           & -0.026 & -0.003 & 61.00 & 0.71          & 0.031 \\
        \textbf{LinearRegression}  &  23.854 &  19.464 & 57.00 & 0.62         & -0.480  \\
        \bottomrule
    \end{tabularx}
   \vspace{-35pt}
    \begin{flushleft}
    \end{flushleft}
\end{table*}

\begin{table*}[hb]
\centering
\small
\caption{Graph-based Centrality and Structural Features.}
\label{tab:graph_features}
\begin{tabularx}{\textwidth}{|l|X|}
\hline
\textbf{Feature} & \textbf{Description} \\\hline
Degree Centrality & Measures the number of direct connections a bus stop has with other stops. High values indicate the stop is a major hub with many direct routes passing through it. \\\hline
Closeness Centrality & Reflects how easily a bus stop can reach all other stops in the network. A high closeness value suggests the stop is well-positioned to access the other bus stops efficiently. \\\hline
Betweenness Centrality & Quantifies how often a bus stop lies on the shortest paths between other stops. Stops with high betweenness serve as critical transfer points or bottlenecks. \\\hline
Eigenvector Centrality & Measures the influence of a bus stop within the network, considering both the number and quality of its connections. A stop connected to other well-connected stops will have a high value. \\\hline
Density & Represents the ratio of actual connections to all possible connections in the network. Higher density indicates a more interconnected bus network. \\\hline
Average Clustering Coefficient & Measures the tendency of bus stops to form tightly connected clusters. Higher values suggest stops are more likely to form local groups of interconnected routes. \\\hline
\end{tabularx}
\end{table*}

\begin{table*}[ht]
\centering
\small
\caption{Urban facility features used in the analysis.}
\label{tab:urban_facilities}
\begin{tabularx}{\textwidth}{|l|X|}
\hline
\textbf{Feature} & \textbf{Type of Urban Facility} \\\hline
education & Educational institutions (e.g., schools, colleges). \\\hline
sport & Sports facilities (e.g., gyms, sports halls). \\\hline
playground & Public playgrounds. \\\hline
community\_center & Large community centers. \\\hline
health & Clinics and small health facilities. \\\hline
elderly\_social\_club & Facilities for the elderly. \\\hline
daycares & Daycare centers. \\\hline
Hospital & Hospitals (large-scale health facilities). \\\hline
University & Universities and higher education institutions. \\\hline
Organized\_Shopping\_Center & Large shopping centers. \\\hline
Organized\_Commerce\_Centers & Large commerce centers. \\\hline
Industrial\_Area & Industrial areas/zones. \\\hline
Market & Public markets. \\\hline
Neighborhood\_Shopping\_Centers & Neighborhood-level shopping centers. \\\hline
Old\_City & The historic city center. \\\hline
High-Tech\_Park & High-tech and innovation parks. \\\hline
Civic\_Center & Civic/governmental centers. \\\hline
Street\_Oriented\_Commerce & Commerce along main streets. \\\hline
Street\_Accompanied\_Commerce & Secondary street commerce. \\\hline
\end{tabularx}
\end{table*}

\begin{table*}[ht]
    \centering
    \caption{Wilcoxon Signed-Rank Test (Polygon, ID vs.\ Non-ID): MAE Differences Across 14 Matched Test Sets}
    \textbf{Note}: Negative differences indicate lower MAE with ID features. Statistically significant $p$-values ($<0.05$) are bolded.
    \label{tab:wilcoxon_polygon_id_vs_noid_compact}
    \renewcommand{\arraystretch}{1.1}
    \begin{tabularx}{\textwidth}{l *{4}{>{\centering\arraybackslash}X}}
        \toprule
        \textbf{Model} &
        \shortstack{Mean \\ Diff (MAE)} &
        \shortstack{Median \\ Diff (MAE)} &
        \shortstack{Wilcoxon \\ Stat} &
        \shortstack{$p$-\\Value} \\
        \midrule
        \textbf{LightGBM}         & -0.02 & -0.02 & 7.00  & \textbf{0.00} \\
        \textbf{XGBoost}          & -0.03 & -0.02 & 23.00 & \textbf{0.03} \\
        \textbf{LinearRegression} & -5.12 & -0.03 & 42.00 & 0.27 \\
        \textbf{RandomForest}     & -0.00 & -0.00 & 47.00 & 0.38 \\
        \textbf{CatBoost}         &  0.01 &  0.00 & 62.00 & 0.73 \\
        \bottomrule
    \end{tabularx}
\end{table*}

\begin{table*}[ht]
    \centering
    \caption{Wilcoxon Signed-Rank Test (Global, ID vs.\ Non-ID): MAE Differences Across 14 Matched Test Sets}
    \textbf{Note}: Negative differences indicate lower MAE with ID features. Statistically significant $p$-values ($<0.05$) are bolded. 
    \label{tab:wilcoxon_global_id_vs_noid_compact}
    \renewcommand{\arraystretch}{1.2}
    \begin{tabularx}{\textwidth}{l *{4}{>{\centering\arraybackslash}X}}
        \toprule
        \textbf{Model} &
        \shortstack{Mean \\ Diff (MAE)} &
        \shortstack{Median \\ Diff (MAE)} &
        \shortstack{Wilcoxon \\ Stat} &
        \shortstack{$p$-\\Value} \\
        \midrule
        \textbf{LightGBM}         & -0.04 & -0.05 & 8.00  & \textbf{0.00} \\
        \textbf{CatBoost}         & -0.02 & -0.01 & 34.00 & 0.13 \\
        \textbf{RandomForest}     & -0.00 & -0.00 & 47.00 & 0.38 \\
        \textbf{XGBoost}          &  0.01 & -0.01 & 44.00 & 0.31 \\
        \textbf{LinearRegression} & -32.80 & -16.00 & 2.00 & \textbf{0.00} \\
        \bottomrule
    \end{tabularx}
\end{table*}

\begin{figure*}[hb]
     \centering
     \includegraphics[width=1\linewidth]{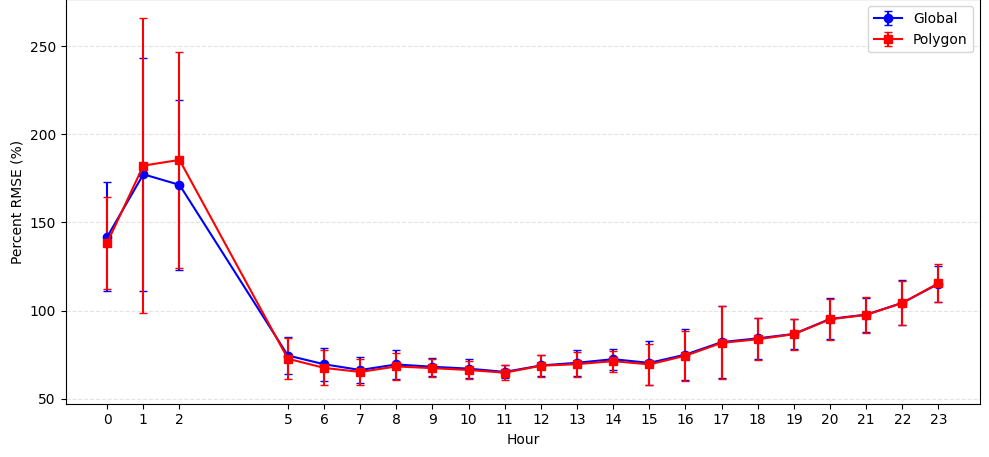}
     \caption{Percent RMSE of LightGBM by hour (mean ± std) for global and polygon strategies, showing highest errors during night and early morning hours and lower, more stable errors throughout the day.}
     \label{fig:lightgbm_RMSE_p_hourly}
\end{figure*}

\begin{figure*}[ht]
     \centering
     \includegraphics[width=1\linewidth, height = 0.45\textheight]{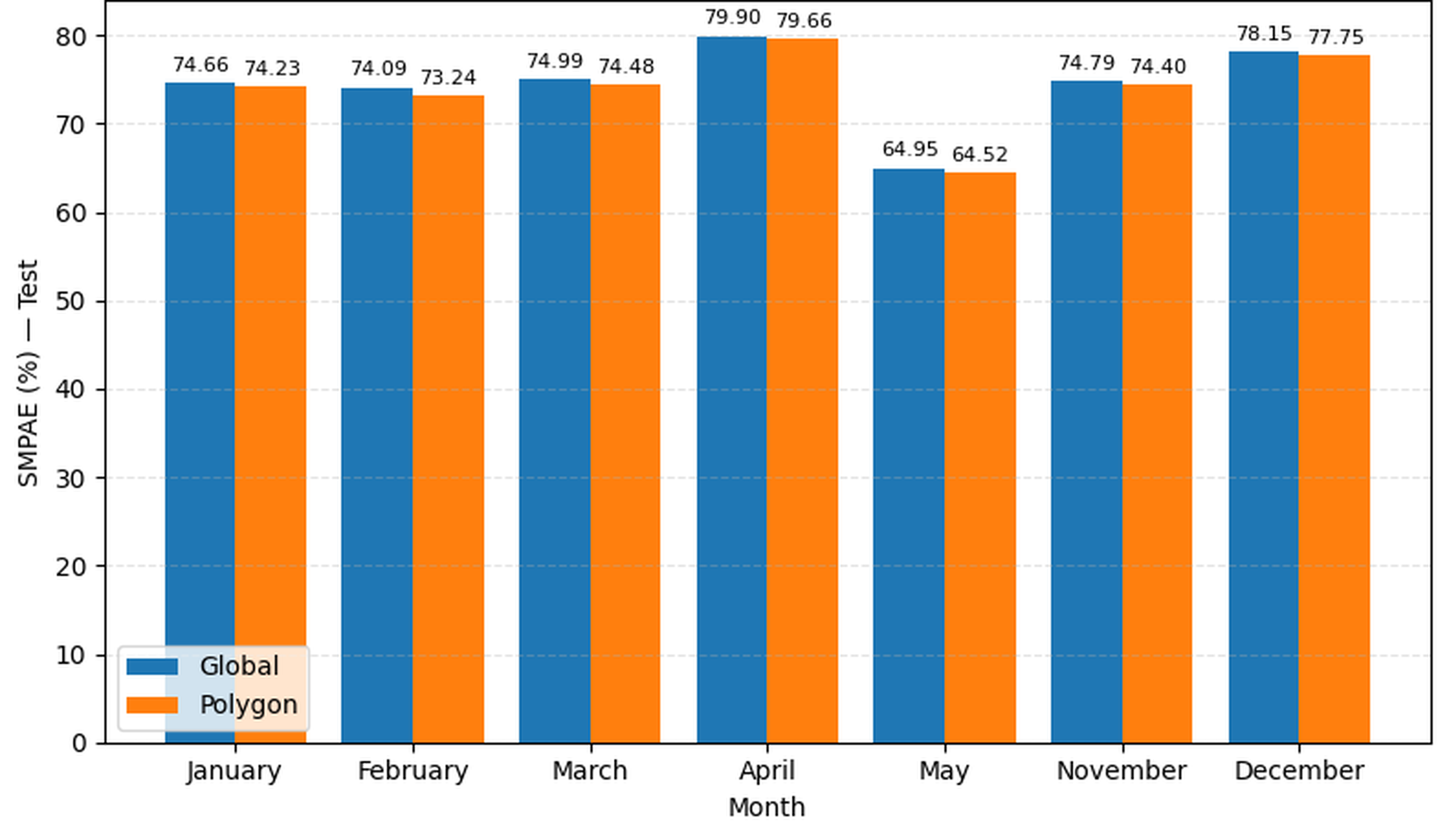}
     \caption{Average monthly sMPAE values of global and polygon-wise LightGBM models across all test sets.}
     \label{fig:lightgbm_month_sMPAE}
\end{figure*}

\begin{figure*}[ht]
     \centering
     \includegraphics[width=1\linewidth]{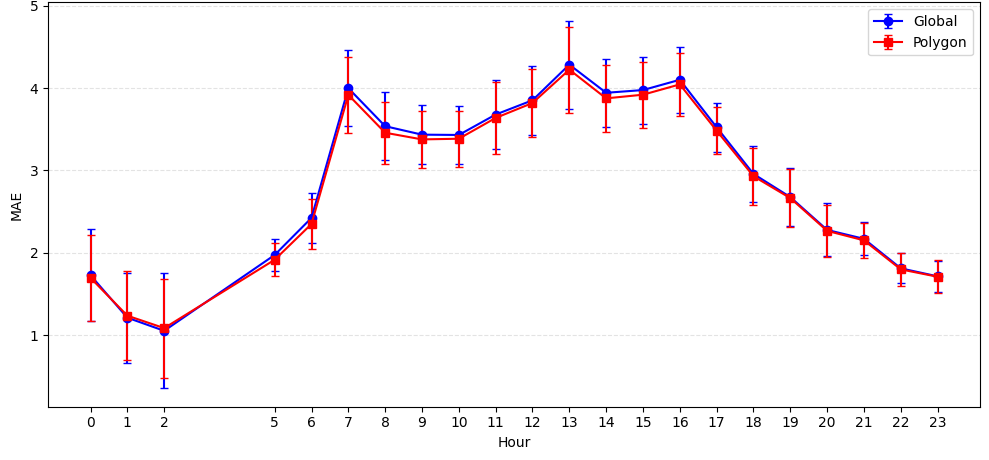}
     \caption{MAE of LightGBM by hour (mean ± std) for global and polygon strategies, showing lower errors during night and early morning hours, higher errors during peak periods, and a gradual decline in the evening.}
     \label{fig:lightGBM_MAE_hourly}
\end{figure*}

\begin{figure*}[ht]
     \centering
     \includegraphics[width=1\linewidth, height = 0.35\textheight]{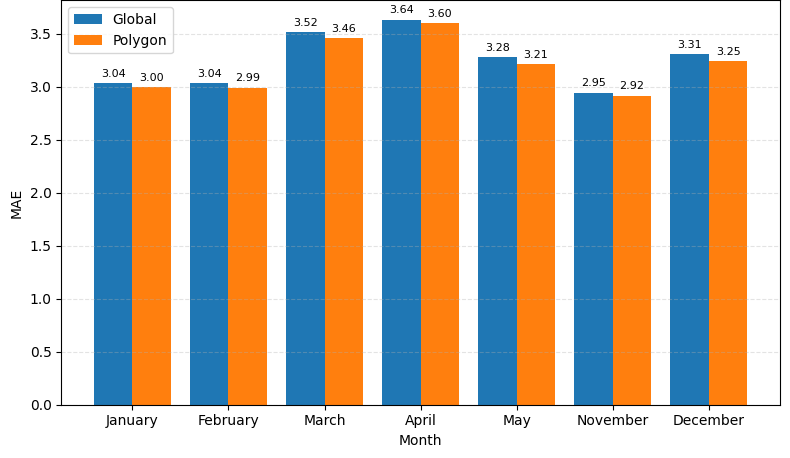}
     \caption{Average monthly MAE values of global and polygon-wise LightGBM models across all test sets.}
     \label{fig:lightGBM_MAE_month}
\end{figure*}

\begin{figure*}[ht]
     \centering
     \includegraphics[width = 0.9\linewidth, height = 0.35\textheight]{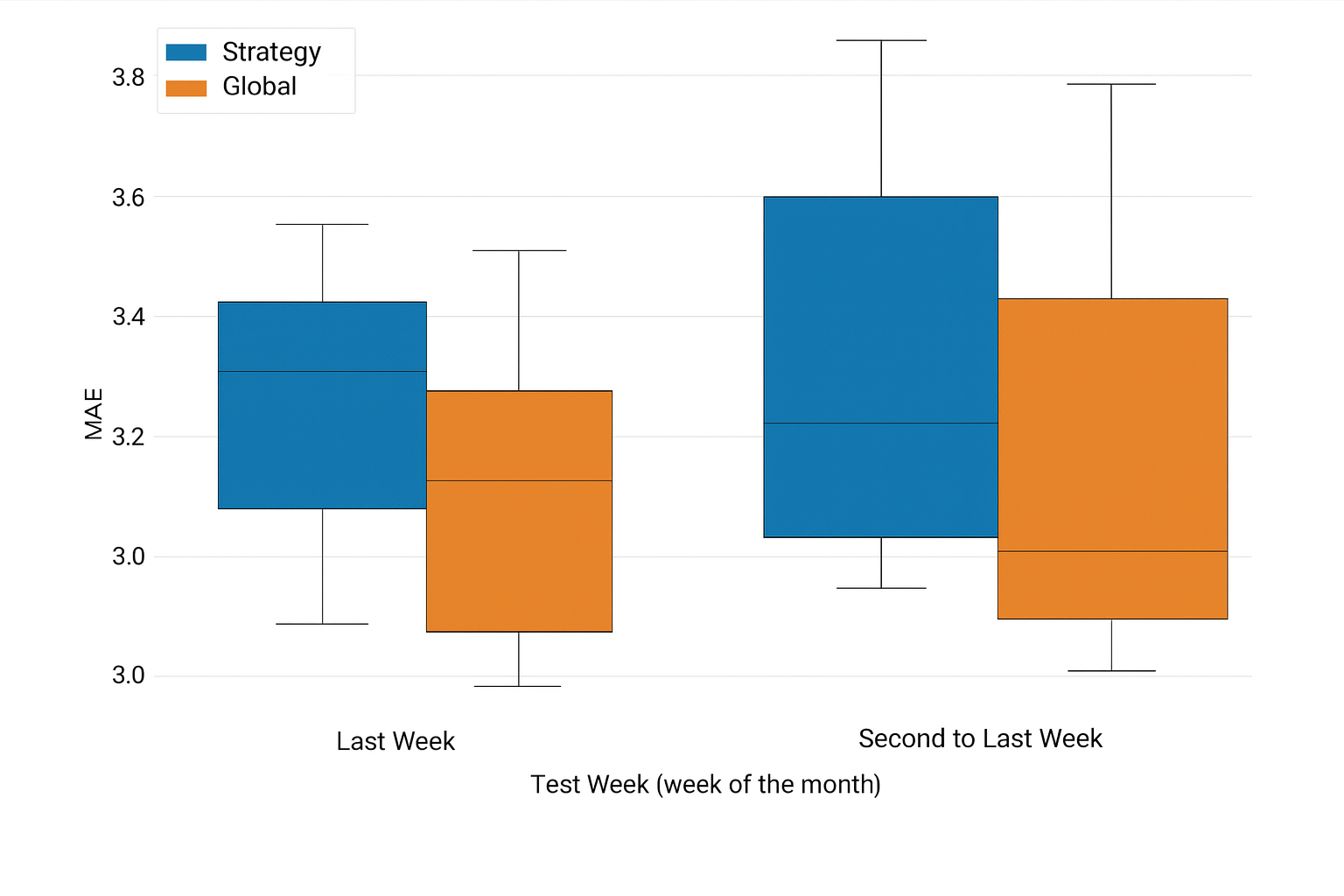}
     \caption{Distribution of LightGBM test MAE across the last week and second to last week of the month, comparing global and polygon strategies. Both strategies show similar error ranges, with the polygon approach generally exhibiting slightly lower central values.}
     \label{fig:lightgmb_MAE_week}
\end{figure*}


\begin{figure}[ht] 
    \centering
    
    \begin{minipage}{0.48\linewidth}
        \centering
        \includegraphics[width=\linewidth]{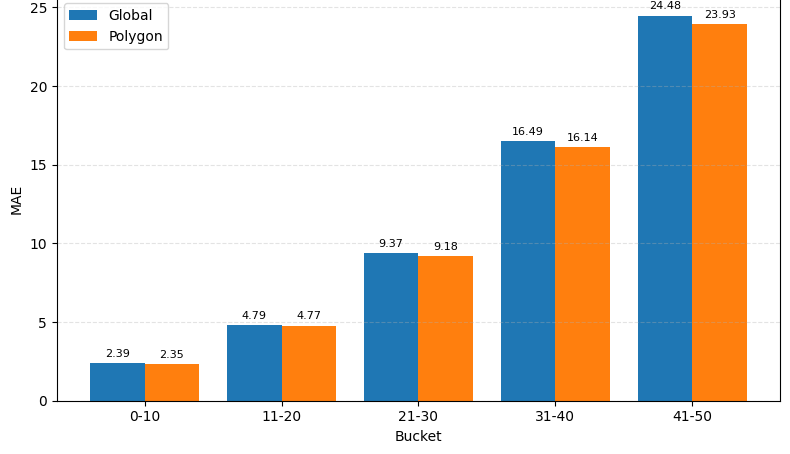}
        \caption{MAE of LightGBM across ridership buckets (0–10 to 41–50) over all experiments, comparing global and polygon strategies. Both approaches show similar performance, with the highest errors in the lowest ridership bucket and a monotone increase in error from 11–20 up to 41–50 ridership.}
        \label{fig:lightGBM_MAE_buckets}
    \end{minipage}\hfill 
    \begin{minipage}{0.48\linewidth}
        \centering
        \includegraphics[width=\linewidth]{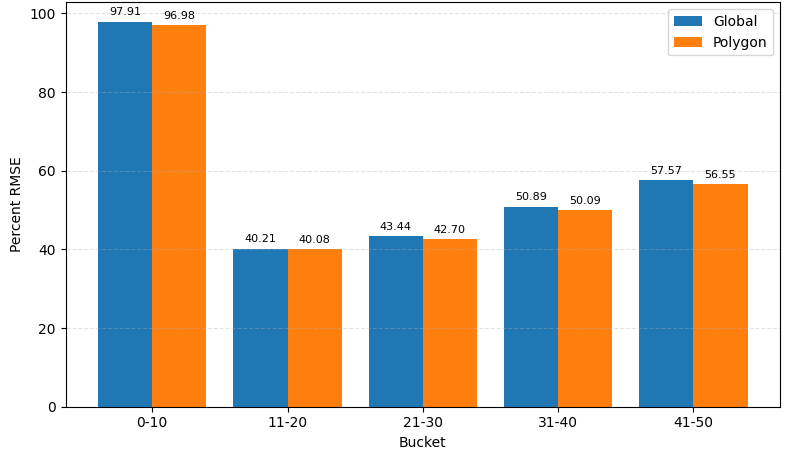}
        \caption{RMSE of LightGBM across ridership buckets (0–10 to 41–50) over all experiments, comparing global and polygon strategies. Both approaches show similar performance, with the highest errors in the lowest ridership bucket and a monotone increase in error from 11–20 up to 41–50 ridership.}
        \label{fig:lightGBM_RMSE_buckets}
    \end{minipage}

\end{figure}
\begin{figure*}[ht]
     \centering
     \includegraphics[width=1\linewidth]{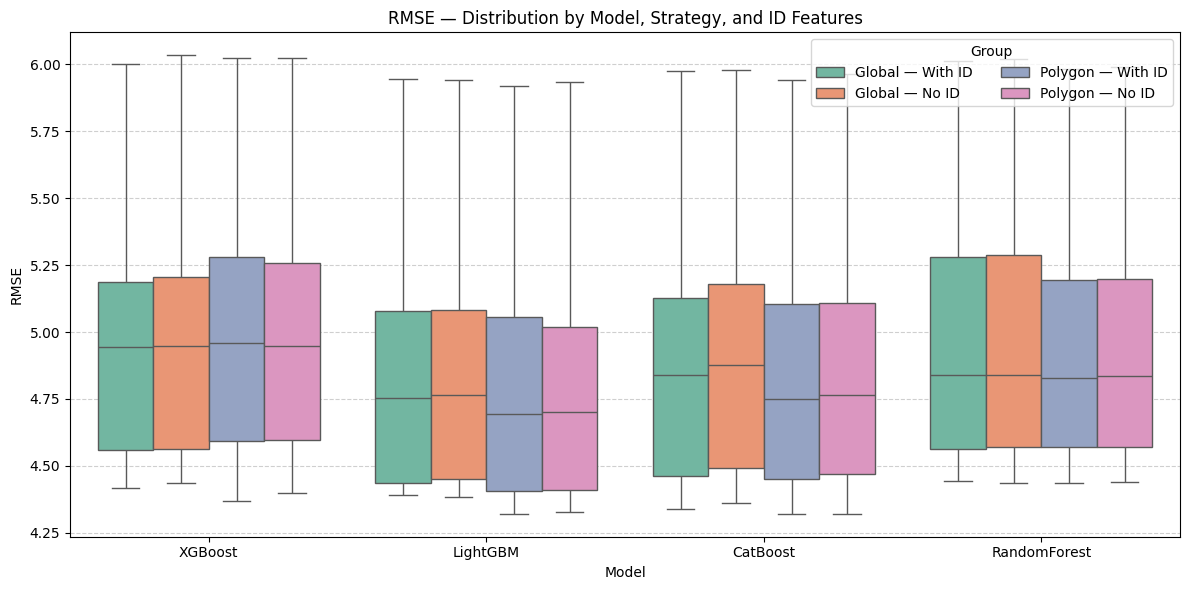}
     \caption{Distribution of RMSE values across models (XGBoost, LightGBM, CatBoost, Random Forest) under global vs. polygon strategies and with vs. without ID features}
    \label{fig:RMSE_global_polygon_id_no_id}
\end{figure*}

\begin{figure*}[t]
  \centering
  \begin{subfigure}{0.49\textwidth}
    \centering
    \includegraphics[trim=10mm 5mm 10mm 5mm, clip, width=\linewidth, height=0.5\textheight, keepaspectratio]{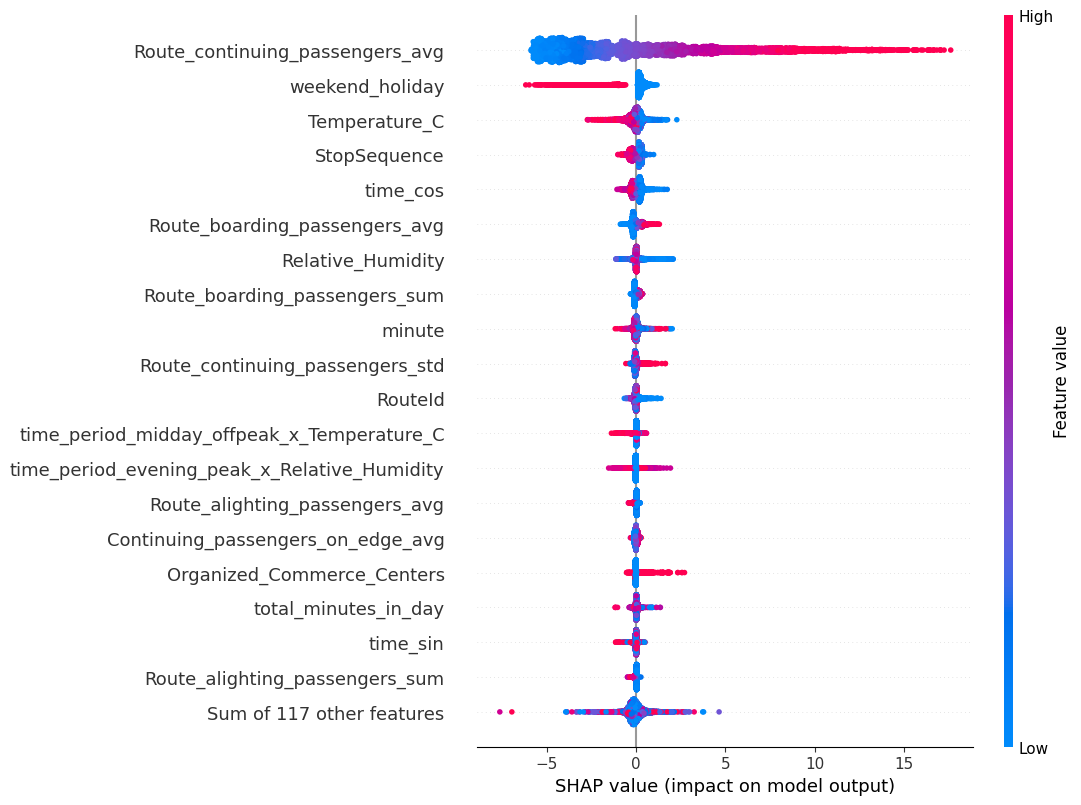}
    \caption{May with identifier features}
    \label{fig:shap_may_id}
  \end{subfigure}
  \hfill
  \begin{subfigure}{0.49\textwidth}
    \centering
    \includegraphics[trim=10mm 5mm 10mm 5mm, clip, width=\linewidth, height=0.5\textheight, keepaspectratio]{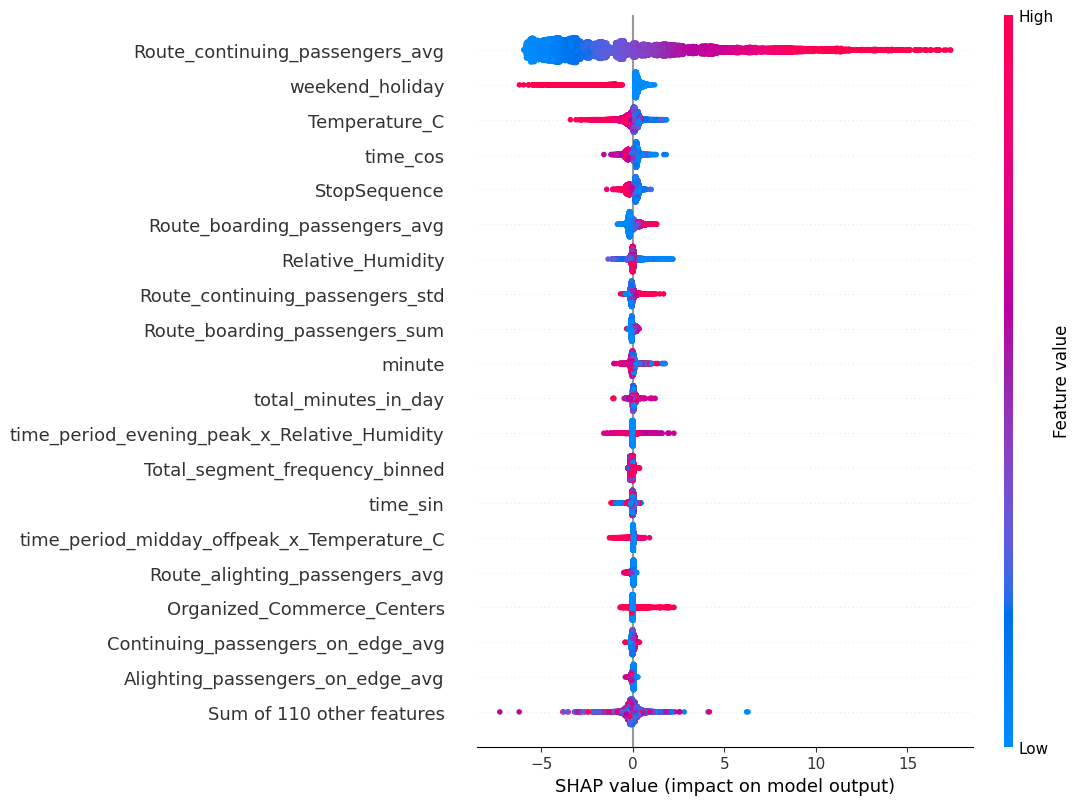}
    \caption{May without identifier features}
    \label{fig:shap_may_noid}
  \end{subfigure}

  \caption{SHAP values for May (polygon-level LightGBM).}
  \label{fig:shap_may}
\end{figure*}

\onecolumn
\begin{center}

\small
\renewcommand{\arraystretch}{1.1}

\begin{longtable}{|p{5cm}|p{8cm}|}

\caption{Base features used in the modeling framework.}

\label{tab:features_base} \\

\hline

\textbf{Feature} & \textbf{Description} \\

\hline

\endfirsthead

\hline

\textbf{Feature} & \textbf{Description} \\

\hline

\endhead
Route Id & Route identifier of the bus line. \\\hline

Direction & Travel direction of the route. \\\hline

Alternative & Alternative variant of the route. \\\hline

Stop Code & Unique identifier of the bus stop. \\\hline

Stop Sequence & Sequential order of the stop along the trip. \\\hline

Passengers Continue & Number of passengers continuing at the stop. \\\hline

Lat & Latitude of the bus stop. \\\hline

Long & Longitude of the bus stop. \\\hline

Education & Presence of educational institutions near the stop. \\\hline

Sport & Presence of sports facilities near the stop. \\\hline

Playground & Presence of playgrounds near the stop. \\\hline

Community Center & Presence of community centers near the stop. \\\hline

Health & Presence of health facilities near the stop. \\\hline

Elderly Social Club & Presence of elderly social clubs near the stop. \\\hline

Daycares & Presence of daycare centers near the stop. \\\hline

Hospital & Presence of hospitals near the stop. \\\hline

University & Presence of universities near the stop. \\\hline

Organized Shopping Center & Presence of organized shopping centers. \\\hline

Organized Commerce Centers & Presence of organized commerce centers. \\\hline

Industrial Area & Presence of industrial areas. \\\hline

Market & Presence of markets near the stop. \\\hline

Neighborhood Shopping Centers & Presence of neighborhood shopping centers. \\\hline

Old City & Indicator for location in the old city. \\\hline

High-Tech Park & Presence of high-tech employment zones. \\\hline

Civic Center & Presence of civic centers. \\\hline

Street Oriented Commerce & Presence of street-oriented commerce. \\\hline

Street Accompanied Commerce & Presence of commerce accompanied by residential streets. \\\hline

Relative Humidity (\%) & Relative humidity at the departure time. \\\hline

Temperature (°C) & Temperature at the departure time. \\\hline

Rain Amount (mm) & Rainfall amount at the departure time. \\\hline

Hour & Hour of departure. \\\hline

Minute & Minute of departure. \\\hline

Total Minutes in Day & Total minutes since midnight. \\\hline

Time Sin & Cyclical time encoding (sine). \\\hline

Time Cos & Cyclical time encoding (cosine). \\\hline

Time Period Early Morning/Night & Indicator for early morning/late night. \\\hline

Time Period Evening Peak & Indicator for evening peak hours. \\\hline

Time Period Midday Offpeak & Indicator for midday off-peak. \\\hline

Time Period Morning Peak & Indicator for morning peak hours. \\\hline

Time Period Night Time & Indicator for night time. \\\hline

Weekend/Holiday & Indicator for weekends/holidays. \\\hline

Previous Station & Previous station before the current stop. \\\hline

Destination & Destination station of the trip. \\\hline

Social Economic Score & Socioeconomic score of the stop’s neighborhood. \\\hline

\end{longtable}
\end{center}

\begin{center}

\small

\begin{longtable}{|p{5cm}|p{9cm}|}

\caption{Centrality features.}

\label{tab:features_centrality} \\

\hline

\textbf{Feature} & \textbf{Description} \\

\hline

\endfirsthead

\hline

\textbf{Feature} & \textbf{Description} \\

\hline

\endhead

Median betweenness centrality & Median betweenness centrality across stops/segments. \\\hline

Mean betweenness centrality & Average betweenness centrality across stops/segments. \\\hline

Standard deviation of betweenness centrality & Variability of betweenness centrality. \\\hline

Median closeness centrality & Median closeness centrality across stops. \\\hline

Mean closeness centrality & Average closeness centrality. \\\hline

Standard deviation of closeness centrality & Variability of closeness centrality. \\\hline

Median eigenvector centrality & Median eigenvector centrality across stops. \\\hline

Mean eigenvector centrality & Average eigenvector centrality. \\\hline

Standard deviation of eigenvector centrality & Variability of eigenvector centrality. \\\hline

Network density & Proportion of realized links out of all possible links. \\\hline

Average clustering coefficient & Local cohesion of the network (average across stops). \\\hline

In-degree centrality (station-level) & Number of incoming links to a stop. \\\hline

Out-degree centrality (station-level) & Number of outgoing links from a stop. \\\hline

Betweenness centrality (station-level) & Share of shortest paths that pass through a stop. \\\hline

Closeness centrality (station-level) & Proximity of a stop to all other stops. \\\hline

Eigenvector centrality (station-level) & Influence of a stop based on its neighbors’ importance. \\\hline

\end{longtable}

\end{center}

\begin{center}

\small

\renewcommand{\arraystretch}{1.1}

\begin{longtable}{|p{0.3\textwidth}|p{0.55\textwidth}|}

\caption{Edge features.}

\label{tab:features_edge} \\

\hline

\textbf{Feature} & \textbf{Description} \\

\hline

\endfirsthead

\hline

\textbf{Feature} & \textbf{Description} \\

\hline

\endhead

Edge betweenness (unweighted) & Importance of the segment in connecting different parts of the network. \\\hline

Number of routes per edge & Number of distinct routes passing through the segment. \\\hline

Edge distance & Length of the segment in meters. \\\hline

Boarding passengers on edge (sum) & Total boarding passengers on the segment. \\\hline

Boarding passengers on edge (avg) & Average boarding passengers on the segment. \\\hline

Boarding passengers on edge (std) & Variability of boarding passengers on the segment. \\\hline

Alighting passengers on edge (sum) & Total alighting passengers on the segment. \\\hline

Alighting passengers on edge (avg) & Average alighting passengers on the segment. \\\hline

Alighting passengers on edge (std) & Variability of alighting passengers on the segment. \\\hline

Continuing passengers on edge (sum) & Total continuing passengers on the segment. \\\hline

Continuing passengers on edge (avg) & Average continuing passengers on the segment. \\\hline

Continuing passengers on edge (std) & Variability of continuing passengers on the segment. \\\hline

Segment frequency (binned, edge-level) & Frequency of trips over the edge (binned). \\\hline

\end{longtable}

\end{center}

\begin{center}

\small

\renewcommand{\arraystretch}{1.1}

\begin{longtable}{|p{0.27\textwidth}|p{0.5\textwidth}|}

\caption{Weight statistics features.}

\label{tab:features_weights} \\

\hline

\textbf{Feature} & \textbf{Description} \\

\hline

\endfirsthead

\hline

\textbf{Feature} & \textbf{Description} \\

\hline

\endhead

Mean continuing passengers & Average number of continuing passengers across segments. \\\hline

Median continuing passengers & Median number of continuing passengers. \\\hline

Std. dev. continuing passengers & Variability of continuing passengers. \\\hline

Min continuing passengers & Minimum continuing passengers observed. \\\hline

Max continuing passengers & Maximum continuing passengers observed. \\\hline

Total continuing passengers & Sum of continuing passengers. \\\hline

Count of continuing passenger obs. & Number of segment-level continuing passenger values. \\\hline

Mean alighting passengers & Average number of alighting passengers. \\\hline

Median alighting passengers & Median number of alighting passengers. \\\hline

Std. dev. alighting passengers & Variability of alighting passengers. \\\hline

Min alighting passengers & Minimum alighting passengers observed. \\\hline

Max alighting passengers & Maximum alighting passengers observed. \\\hline

Total alighting passengers & Sum of alighting passengers. \\\hline

Count of alighting passenger obs. & Number of segment-level alighting passenger values. \\\hline

Mean boarding passengers & Average number of boarding passengers. \\\hline

Median boarding passengers & Median number of boarding passengers. \\\hline

Std. dev. boarding passengers & Variability of boarding passengers. \\\hline

Min boarding passengers & Minimum boarding passengers observed. \\\hline

Max boarding passengers & Maximum boarding passengers observed. \\\hline

Total boarding passengers & Sum of boarding passengers. \\\hline

Count of boarding passenger obs. & Number of segment-level boarding passenger values. \\\hline

Mean segment frequency (binned) & Average frequency of trips per segment (binned). \\\hline

Median segment frequency (binned) & Median segment frequency (binned). \\\hline

Std. dev. segment frequency (binned) & Variability of binned segment frequency. \\\hline

Min segment frequency (binned) & Minimum segment frequency (binned). \\\hline

Max segment frequency (binned) & Maximum segment frequency (binned). \\\hline

Total segment frequency (binned) & Sum of binned segment frequencies. \\\hline

Count of segment frequency obs. (binned) & Number of frequency observations per segment (binned). \\\hline

Mean distance between stops & Average inter-stop distance. \\\hline

Median distance between stops & Median inter-stop distance. \\\hline

Std. dev. distance between stops & Variability of inter-stop distance. \\\hline

Min distance between stops & Minimum inter-stop distance. \\\hline

Max distance between stops & Maximum inter-stop distance. \\\hline

Total distance between stops & Sum of all inter-stop distances. \\\hline

Count of distance observations & Number of segment-level distance values. \\\hline

\end{longtable}

\end{center}

\begin{center}

\small

\renewcommand{\arraystretch}{1.1}

\begin{longtable}{|p{0.3\textwidth}|p{0.40\textwidth}|}

\caption{Route-specific aggregated features.}

\label{tab:features_route} \\

\hline

\textbf{Feature} & \textbf{Description} \\

\hline

\endfirsthead

\hline

\textbf{Feature} & \textbf{Description} \\

\hline

\endhead

Route segment frequency & Frequency of segments within a route. \\\hline

Route boarding passengers (sum) & Total boardings aggregated at route level. \\\hline

Route boarding passengers (avg) & Average boardings per route. \\\hline

Route boarding passengers (std) & Variability of boardings at route level. \\\hline

Route alighting passengers (sum) & Total alightings aggregated at route level. \\\hline

Route alighting passengers (avg) & Average alightings per route. \\\hline

Route alighting passengers (std) & Variability of alightings at route level. \\\hline

Route continuing passengers (sum) & Total continuing passengers at route level. \\\hline

Route continuing passengers (avg) & Average continuing passengers per route. \\\hline

Route continuing passengers (std) & Variability of continuing passengers at route level. \\\hline

Average route distance & Mean inter-stop distance across the route. \\\hline

Route segment frequency (binned) & Frequency of segments within the route (binned). \\\hline

Total segment frequency (binned, route-level) & Total binned frequency aggregated at route level. \\\hline

\end{longtable}

\end{center}

\begin{figure*}[p]
  \centering
  \includegraphics[width=0.7\textwidth]{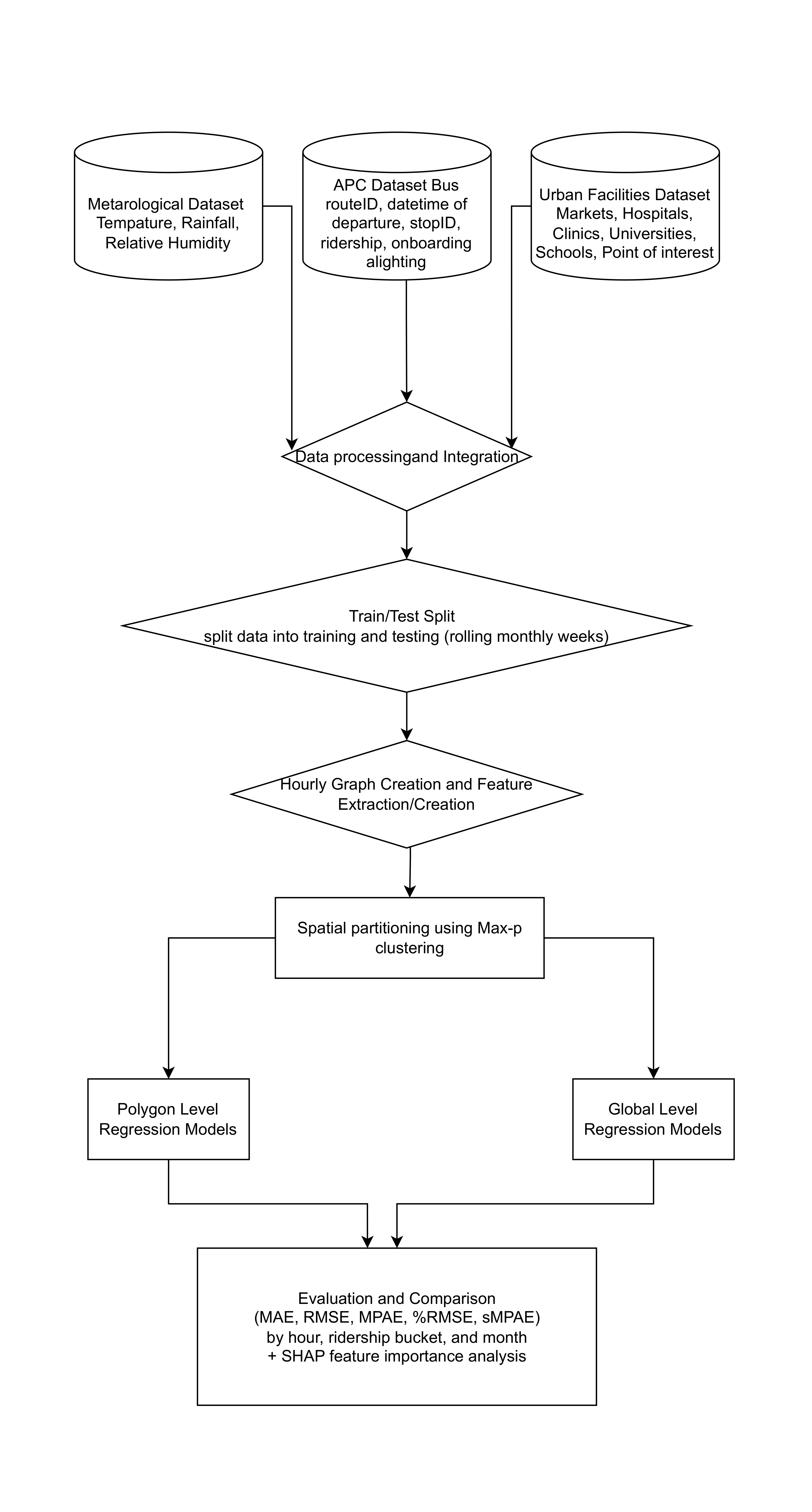} 
  \caption{Experiments workflow}
  \label{fig:workflow}
\end{figure*}


\end{document}